\documentclass{article}
\usepackage{arxiv}
\usepackage[utf8]{inputenc}
\usepackage[T1]{fontenc}  
\usepackage{times}
\usepackage{float}
\usepackage{hyperref}
\usepackage{subcaption}
\usepackage{graphicx}
\usepackage{adjustbox}
\usepackage{multirow}
\usepackage{pbox}
\usepackage[font=small]{caption}
\usepackage{amsmath}
\usepackage{amssymb}

\usepackage{algorithm}
 \usepackage{algpseudocode}
\usepackage{pgfplots}
\newcommand{\bx}{{\boldsymbol{x}}}
\newcommand{\bu}{{\boldsymbol{u}}}

\usepgfplotslibrary{groupplots,dateplot}
\usetikzlibrary{patterns,shapes.arrows}
\pgfplotsset{compat=newest}

\usepackage{adjustbox}
 \usepackage{multirow}
\usepackage{pbox}
\usepackage{tikz}
\usetikzlibrary{shapes, arrows.meta, positioning}

\usepgfplotslibrary{groupplots}
\usepgfplotslibrary{fillbetween}
\usetikzlibrary{arrows,decorations.pathmorphing,positioning,fit,trees,shapes,shadows,automata,calc} 
\usetikzlibrary{patterns,arrows,arrows.meta,calc,shapes,shadows,decorations.pathmorphing,decorations.pathreplacing,automata,shapes.multipart,positioning,shapes.geometric,fit,circuits,trees,shapes.gates.logic.US,fit, matrix,arrows.meta, quotes}
\usetikzlibrary{backgrounds,scopes}
\usepackage{ulem}

\usepackage{tabu}

\usepackage{xspace}

\usepackage{pgfplots}
\usepgfplotslibrary{colorbrewer}
\pgfplotsset{compat=1.17} 
\newcommand{\ours}{TransformerMPPI\xspace}

\usetikzlibrary{pgfplots.statistics, pgfplots.colorbrewer} 
\usepackage{pgfplotstable}
\usepackage{filecontents}

\usepackage{makecell} 

\usepackage[numbers]{natbib} 

\title{Transformer-based Model Predictive Path Integral Control}

\author{%
  {Shrenik Zinage$\textsuperscript{*}$} \\
  School of Mechanical Engineering\\
  Purdue University\\
  West Lafayette, IN \\
  \texttt{szinage@purdue.edu} \\
  \And
  {Vrushabh Zinage$\textsuperscript{*}$} \\
  Department of Aerospace Engineering \\
  University of Texas at Austin\\
  Austin, TX\\
  \texttt{vrushabh.zinage@austin.utexas.edu} \\
  \And
  {Efstathios Bakolas} \\
  Department of Aerospace Engineering \\
  University of Texas at Austin\\
  Austin, TX\\
  \texttt{bakolas@austin.utexas.edu} \\
}

\renewcommand{\shorttitle}{Transformer MPPI}
\hypersetup{
  pdftitle={Transformer MPPI},
  pdfauthor={Shrenik Zinage, Vrushabh Zinage, Efstathios Bakolas}
}

\begin{document}
\maketitle

\footnotetext[1]{\textsuperscript{*}Both authors contributed equally to this research.}

\begin{abstract}
This paper presents a novel approach to improve the Model Predictive Path Integral (MPPI) control by using a transformer to initialize the mean control sequence. Traditional MPPI methods often struggle with sample efficiency and computational costs due to suboptimal initial rollouts. We propose TransformerMPPI, which uses a transformer trained on historical control data to generate informed initial mean control sequences. \ours combines the strengths of the attention mechanism in transformers and sampling-based control, leading to improved computational performance and sample efficiency. The ability of the transformer to capture long-horizon patterns in optimal control sequences allows \ours to start from a more informed control sequence, reducing the number of samples required, and accelerating convergence to optimal control sequence. We evaluate our method on various control tasks, including avoidance of collisions in a 2D environment and autonomous racing in the presence of static and dynamic obstacles. Numerical simulations demonstrate that \ours consistently outperforms traditional MPPI algorithms in terms of overall average cost, sample efficiency, and computational speed in the presence of static and dynamic obstacles.
\end{abstract}


\section{Introduction}
\label{sec:intro}
Model Predictive Control (MPC) \cite{kouvaritakis2016model_mpc1,schwenzer2021review_mpc2,zinage2024transformermpc} has emerged as a powerful framework for optimal control across a diverse range of applications, including robotics \cite{bangura2014real_mpc3,zanelli2018nonlinear_mpc4}, space \cite{hegrenaes2005spacecraft} and ocean \cite{zinage2020comparative, zinage2021comparative} operations.
At its core, MPC involves solving a finite-horizon optimal control problem at each time step, executing the first control input, and then repeating this process in a receding horizon manner. This approach allows MPC to handle nonlinear dynamics, constraints, and objectives while adapting to changing conditions in real-time.

Within the broader optimization-based control paradigm, sampling-based control methods, particularly Model Predictive Path Integral (MPPI) control have gained significant traction in recent years, particularly in domains characterized by highly nonlinear dynamics, nonconvex cost functions, or learned models. MPPI reformulates the optimal control problem from an information-theoretic perspective, drawing connections to free energy and the Feynman-Kac lemma. This formulation leads to an update rule for the control distribution, where the sampled trajectories are weighted exponentially according to their costs. MPPI has demonstrated remarkable empirical success in a variety of challenging control tasks, from agile autonomous driving \cite{mohamed2022autonomous_mppi_2,testouri2023towards_mppi_driving_2} to quadrotor control \cite{mohamed2020model_mppi_quadrotor_1,minarik2024model_mppi_quadrotor_2,lu2022real_mppi_quadrotor_3,jin2023safety_mppi_quadrotor_4} and robotic manipulation \cite{zhang2024multi_mppi_manipulation_1,abraham2020model_manipulation_2,arruda2017uncertainty_manipulation_3,cong2020self_manipulation_4,carius2022constrained_manipulation_5}.
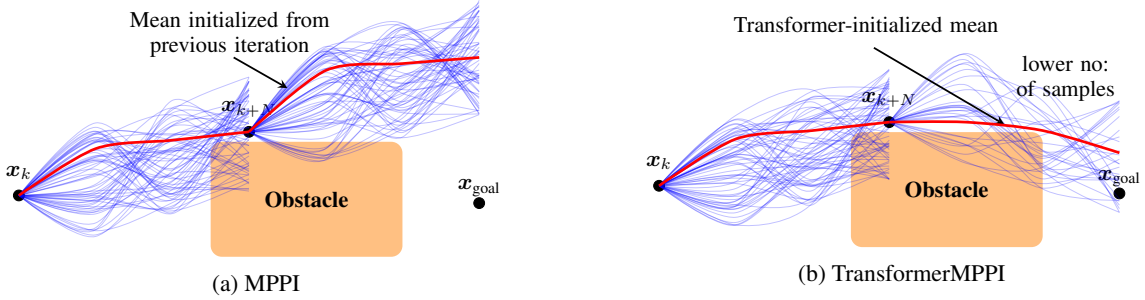
\begin{figure}
    \centering
        \begin{minipage}{0.48\textwidth}
        \centering
        \scalebox{0.85}{
        \begin{tikzpicture}[scale=1.2, >=stealth]

\fill[orange!70, rounded corners=5pt, opacity=0.7] (2.5, -0.6) rectangle (5.0, 0.9);
\node at (3.75, 0.15) {\textbf{Obstacle}};
\node at (0, 0.5) {$\bx_k$};
\node at (3, 1.33) {$\bx_{k+N}$};
\node at (6, 0.3) {$\bx_{\text{goal}}$};
\filldraw[black] (6, 0.1) circle (2pt) node[anchor=south west] {};
\filldraw[black] (0, 0.2) circle (2pt) node[anchor=south west] {};
\filldraw[black] (3, 1.03) circle (2pt) node[anchor=south west] {};
\foreach \i in {1,...,65} {
    \draw[opacity=0.25, color=blue] plot[smooth]
    coordinates {(0, 0.2) 
                 (1, 0.4 + 0.75*rand) 
                 (2, 0.7 + 0.75*rand) 
                 (3, 1.0 + 0.75*rand)
                  };
}

\foreach \i in {1,...,65} {
    \draw[opacity=0.25, color=blue] plot[smooth]
    coordinates {(3, 1.03) 
                 (4, 1.4 + 0.75*rand) 
                 (5, 1.7 + 0.75*rand) 
                 (6, 2.0 + 0.75*rand)
                  };
}
\draw[very thick, red] plot[smooth]
coordinates {(0, 0.2) 
             (1, 0.3 + 0.51) 
             (2, 0.4 + 0.52) 
             (3, 0.5 + 0.53)
             }
node[pos=0.8, above right] {};

\draw[very thick, red] plot[smooth]
coordinates {(3, 1.03) 
             (4, 1.3 + 0.53) 
             (5, 1.4 + 0.52) 
             (6, 1.5 + 0.50)
             }
node[pos=0.8, above right] {};

\node[align=center, text width=3.5cm] at (2.75, 2.3) {{Mean initialized from previous iteration}};
\draw[->, thick, black] (2.8, 2.0) -- (3.5, 1.6);




\end{tikzpicture}
        }
        \subcaption{MPPI}
        \label{fig:mppi_sampling}
    \end{minipage}
    \hfill
    \begin{minipage}{0.48\textwidth}
        \centering
        \scalebox{0.85}{
        \begin{tikzpicture}[scale=1.2, >=stealth]

\fill[orange!70, rounded corners=5pt, opacity=0.7] (2.5, -0.6) rectangle (5.0, 0.9);
\node at (3.75, 0.15) {\textbf{Obstacle}};

\node at (0, 0.5) {$\bx_k$};
\node at (6, 0.3) {$\bx_{\text{goal}}$};
\filldraw[black] (6, 0.1) circle (2pt) node[anchor=south west] {};
\node at (3, 1.33) {$\bx_{k+N}$};
\filldraw[black] (0, 0.2) circle (2pt) node[anchor=south west] {};
\filldraw[black] (3, 1.03) circle (2pt) node[anchor=south west] {};
\foreach \i in {1,...,65} {
    \draw[opacity=0.25, color=blue] plot[smooth]
    coordinates {(0, 0.2) 
                 (1, 0.4 + 0.75*rand) 
                 (2, 0.7 + 0.75*rand) 
                 (3, 1.0 + 0.75*rand)
                  };
}

\foreach \i in {1,...,30} {
    \draw[opacity=0.25, color=blue] plot[smooth]
    coordinates {(3, 1.03) 
                 (4, 1.2 + 0.75*rand) 
                 (5, 0.9 + 0.75*rand) 
                 (6, 0.6 + 0.75*rand)
                  };
}

\node[align=center, text width=5.5cm] at (2.7, 2.3) {{Transformer-initialized mean}};
\draw[->, thick, black] (2.8, 2.0) -- (4.5, 1.0);
\node[align=center, text width=2.7cm] at (5.3, 1.6) {{lower no: of samples}};
\draw[very thick, red] plot[smooth]
coordinates {(0, 0.2) 
             (1, 0.3 + 0.51) 
             (2, 0.4 + 0.52) 
             (3, 0.5 + 0.53)
             }
node[pos=0.8, above right] {};

\draw[very thick, red] plot[smooth]
coordinates {(3, 1.03) 
             (4, 1.0 + 0.03) 
             (5, 0.9 + 0.02) 
             (6, 0.6 + 0.03)
             }
node[pos=0.8, above right] {};



\end{tikzpicture}
        }
        \subcaption{\ours}
        \label{fig:transformermppi_sampling}
    \end{minipage}
    \caption{\small Instead of using the mean control sequence from the previous iteration as in the MPPI (Fig. \ref{fig:mppi_sampling}), we leverage an informed transformer initialized mean control sequence. As observed from Fig. \ref{fig:transformermppi_sampling}, if the mean control sequence is initialized via the trained transformer model, it leads to a lower number of samples for \ours to converge to the optimal sequence.}
    \label{fig:mppi_vs_transformer_mppi}
\end{figure}
The main advantages of MPPI over MPC based controllers are that these methods are devoid of traditional nonlinear programming or optimization-based convex solvers in favor of a more flexible zeroth-order optimization approach. By sampling control trajectories from a distribution, evaluating their costs, and updating the sampling distribution based on these evaluations, sampling-based control can handle nonlinear dynamics and constraints without requiring differentiability or convexity of the objective cost function and constraints. In addition, sampling multiple control trajectories can be parallelized on GPUs and are less susceptible to the curse of dimensionality.

MPPI, while promising, can face practical challenges. A key concern is that it may generate infeasible control sequences if all sampled trajectories fall in high-cost regions, potentially violating system constraints or getting stuck in local minima.
To address this, various solutions have been proposed in the literature. 
Reference \cite{kusumoto2019informed} improves MPPI by incorporating an informed sampling process. They use conditional variational autoencoders \cite{kingma2013auto} to learn distributions that emulate samples from a dataset containing optimized controls. These learned distributions are utilized to adjust the mean of sampling distribution prior to executing the MPPI process. Reference \cite{okada2018acceleration} applied optimization techniques originally developed for gradient descent to modify the control input at each time step, thereby improving convergence to the optimal control.
Reference \cite{lambert2020stein} proposed a Bayesian MPC approach that accommodates multimodal distributions and approximates the posterior distribution using Stein variational gradient descent \cite{liu2016stein}. Reference \cite{pravitra2021flying} proposed an iteration step within the MPPI framework, adjusting only the mean of sampling distribution from the weighted average of the previous MPPI iteration. Reference \cite{wang2021variational} used the Tsallis divergence to formulate a sampling-based algorithm and used a similar iteration procedure as in \cite{pravitra2021flying} to refine the distribution within the control update step. However, their approach involves sampling and optimization over different distributions at each control step. On the other hand, MPPI-based approaches that efficiently handle uncertainties have been proposed in the literature. Tube-MPPI \cite{williams2018robust_tube_mppi} incorporates an iterative Linear Quadratic Gaussian (iLQG) controller to track the MPPI-generated trajectory, although it requires dynamics linearization. Reference \cite{pravitra2020_l1_mppi} uses a nonlinear L1 adaptive controller to handle model uncertainty. Reference \cite{yin2022trajectory_mppi_cs_1} focuses on improving MPPI's sampling techniques. One method improves MPPI by incorporating the covariance steering principle. Reference \cite{mohamed2022autonomous_mppi_2} samples from a product of normal and log-normal distributions instead of just a Gaussian distribution. These techniques lead to more efficient trajectories, better state-space exploration, and reduced risk of local minima compared to standard MPPI.

Previous approaches to informed sampling in model predictive control, such as those using conditional variational autoencoders or optimization methods, often suffer from limited generalization and high computational costs. In certain scenarios, they often lack generalizability to problems with time-varying parameters, such as environments with dynamic obstacles. These methods typically struggle to adapt to changing problem conditions, limiting their applicability in real-world scenarios where the environment is constantly evolving. To address this limitation, transformers \cite{vaswani2017attention} offer a more robust solution for informed mean initialization, as their attention mechanisms and ability to process long-range dependencies allow them to better capture and adapt to temporal variations in problem parameters, making them particularly well-suited for handling dynamic environments and obstacles in continuous control tasks.

In this paper, we propose a novel approach to improve the performance and efficiency of MPPI by leveraging transformers for initializing an informed mean control sequence during sampling of the control inputs. Our method, termed \ours, addresses the following key limitations of standard MPPI \cite{williams2018information}. The first is on the initialization quality. As shown in Fig. \ref{fig:mppi_vs_transformer_mppi}, traditional MPPI often relies on zero initialization or initialization from the previous iteration \cite{williams2016aggressive_previous_iteration}, which can lead to suboptimal performance, especially in high-dimensional control spaces. \ours uses a transformer model trained on historical optimal control data to generate close to optimal initial mean control sequences, providing a more informed starting point for optimization. Second, is the sample and computational efficiency. By starting from a better initialization, \ours can potentially reduce the number of samples required to converge to a good solution, improving computational efficiency even in the presence of dynamic obstacles. 

The remainder of this paper is organized as follows. Section \ref{sec:prelim} discusses the preliminaries followed by the proposed approach in Section \ref{sec:proposed_approach}. Section \ref{sec:results} presents the numerical simulations for our approach, followed by the concluding remarks in Section \ref{sec:conclusions}.

\section{Problem Formulation and Preliminaries \label{sec:prelim}}
In this section, we discuss the problem statement and briefly discuss the mathematical formulation of MPPI control.

\subsection{Problem Formulation\label{subsec:problem_formulation}}
Consider the discrete-time nonlinear system as follows:
\begin{align}
    \bx_{k+1}=f\left(\bx_k, \boldsymbol{w}_k\right),\quad \bx_0=\bx^0,\label{eqn:nonlinear_dynamics}
\end{align}
where the state $\bx_k \in \mathbb{R}^{n}$, $f$ is a time-invariant nonlinear state transition function of the system, $\boldsymbol{w}_k\sim \mathcal{N}(\bu_k,\Sigma_\bu)$ is the perturbed control input with mean control $\bu_k\in\mathbb{R}^m$ and covariance $\Sigma_\bu$.
Within a finite time horizon $N$, we denote the control sequence $\bu_{[t,t+N]}= \left[\bu^\mathrm{T}_t, \bu^\mathrm{T}_{t+1}, \ldots, \bu^\mathrm{T}_{t+N-1}\right]^{\mathrm{T}} \in \mathbb{R}^{m N}$ and the corresponding state trajectory $\bx_{[t,t+N+1]}=\left[\bx^\mathrm{T}_t, \bx^\mathrm{T}_{t+1}, \ldots, \bx^\mathrm{T}_{t+N}\right]^{\mathrm{T}} \in \mathbb{R}^{n(N+1)}$. Let $\mathcal{X}^d$ denote the environment space with $\mathcal{X}_{\text{r}}\left(\bx_k\right) \subset \mathcal{X}^d$ and $\mathcal{X}_{\text {obs }} \subset \mathcal{X}^d$ representing the region occupied by the agent and the obstacles, respectively.
The objective of a stochastic optimal control problem is to synthesize an optimal control sequence $\bu_{[t,t+N]}$ at time $t$, that generates a collision-free trajectory, guiding the agent from its current state $\bx_s$ to the desired state $\bx_f$, while minimizing the cost function $J$ subject to specified nonlinear constraints. The uncontrolled $p(W)$ and controlled $q(W)$ probability density functions (PDF) are given respectively by: 
{
\begin{align}
& p(W)= \prod_{i=0}^{N-1}\left((2 \pi)^m|\Sigma_\bu|\right)^{-1 / 2} \exp \left(-\frac{1}{2} \boldsymbol{w}_i^\mathrm{T} \Sigma^{-1}_\bu \boldsymbol{w}_i\right) ,\nonumber\\
\
&q(W)=  \prod_{i=0}^{N-1}\left((2 \pi)^m|\Sigma_\bu|\right)^{-1 / 2} \exp \left(-\frac{1}{2}\left(\boldsymbol{w}_i-\bu_i\right)^\mathrm{T} \Sigma^{-1}_\bu\left(\boldsymbol{w}_i-\bu_i\right)\right),\nonumber
\end{align}
where $W=\left\{\boldsymbol{w}_0, \boldsymbol{w}_1, \cdots, \boldsymbol{w}_{N-1}\right\}$
}
The optimization problem is formulated as:
\begin{subequations}
\begin{align}
   \underset{\mathcal{U}}{\min}\quad & \mathbb{E}_\mathbb{Q}[J(W)]=\mathbb{E}_\mathbb{Q}\left[\phi\left(\bx_N\right)+\sum_{k=0}^{N-1}\left(s\left(\bx_k\right)+\frac{1}{2} \bu_k^{\mathrm{T}} R \bu_k\right)\right]\label{eqn:objective_function} \\
\text { s.t. } \quad & \bx_{k+1}=f\left(\bx_k, \boldsymbol{w}_k\right), \quad \boldsymbol{w}_k \sim \mathcal{N}\left(\bu_k, \Sigma_{\bu}\right)\label{eqn:constraint_1} \\
& \mathcal{X}_{\text{r}}\left(\bx_k\right) \cap \mathcal{X}_{\text{obs}}=\emptyset, \quad h\left(\bx_k, \bu_k\right) \leq 0\label{eqn:constraint_2} \\
& \bx_0=\bx^0, \quad \bu_k \in \mathcal{U},\quad \bx_k \in \mathcal{X}\label{eqn:constraint_3} 
\end{align}
\label{eqn:optimal_control_problem}
\end{subequations}
where $R \in \mathbb{S}^{m \times m}_{+}$ is a positive-definite matrix, $\mathcal{U}\subset \mathbb{R}^m$ denotes the set of constrained control inputs, $h(\bx_k,\bu_k)$ is a set of nonlinear inequality constraints for the agent and $\mathcal{X}$ denotes the set of all possible states $\bx_k$. The terminal cost function $\phi\left(\bx_N\right)$ and the running cost function $s\left(\bx_k\right)$ can be arbitrary functions (that includes non-smooth functions as well), offering a flexible and general approach compared to MPC based controllers.
\subsection{MPPI Algorithm}

{
Define the free energy as
$$
\mathcal{F}\left(J, \mathbb{P}, \bx_0, \lambda\right)=-\lambda \log \left(\mathbb{E}_{\mathbb{P}}\left[\exp \left(-\frac{1}{\lambda} J(W)\right)\right]\right),
$$
where $\mathbb{P}$ represents a base measure characterizing the uncontrolled input distribution with PDF $p(W)$, and $\lambda>0$ is a tuning parameter. It can be shown that:
\begin{align}
\mathcal{F}\left(J, \mathbb{P}, \bx_0, \lambda\right) \leq \mathbb{E}_{\mathbb{Q}}[J(W)]+\lambda \mathbb{K} \mathbb{L}(\mathbb{Q}| | \mathbb{P}),
\label{eqn:free_energy_ineq}
\end{align}
where $\mathbb{Q}$ is probability measure characterizing the controlled input distribution $W$ with PDF $q(W)$, and $\mathbb{KL}(\mathbb{Q} \| \mathbb{P})$ represents the Kullback–Leibler (KL) divergence between the base and controlled measures. This inequality indicates that the free energy acts as a lower bound for the sum of the expected cost under the controlled distribution and a control cost represented by the KL divergence. 
The stochastic optimal control problem \eqref{eqn:optimal_control_problem} can be solved by minimizing the KL-Divergence between controlled and uncontrolled distributions i.e., $\mathbb{K} \mathbb{L}(\mathbb{Q}| | \mathbb{P})$. To that end, we define an optimal control distribution $\mathbb{Q}^\star$ via its Radon-Nikodym derivative with respect to $\mathbb{P}$ as
$$
\frac{d \mathbb{Q}^\star}{d \mathbb{P}}=\frac{\exp \left(-\frac{1}{\lambda} J(W)\right)}{\mathbb{E}_{\mathbb{P}}\left[\exp \left(-\frac{1}{\lambda} J(W)\right)\right]}.
$$
By substituting $\mathbb{Q}$ with $\mathbb{Q}^\star$ in \eqref{eqn:free_energy_ineq}, it can be shown that $\mathbb{Q}^\star$ is indeed an optimal control distribution, as it attains the lower bound in \eqref{eqn:free_energy_ineq}. The objective is then to align our control distribution $\mathbb{Q}$ with the optimal distribution $\mathbb{Q}^\star$ through KL minimization, resulting in the optimal control sequence $U^\star=\underset{  \mathcal{U} }{\operatorname{argmin}}\; \mathbb{K L}\left(\mathbb{Q}^\star \| \mathbb{Q}\right)$.
This approach leads to a new distribution, $\mathbb{Q}^\star$, for importance sampling, defined by the PDF:
$$
q^\star(W)=\frac{1}{\eta} \exp \left(-\frac{1}{\lambda} J(W)\right) p(W),
$$
where $\eta>0$ is a normalizing constant ensuring that the integral of $q^\star(W)$ over the sample space equals one. The sensitivity parameter $\eta>0$ adjusts the importance of cost differences between trajectories. Consequently, the optimal control input $\bu_i^\star$ is given by: 
$$
\bu_i^\star=\int q^\star(W) \boldsymbol{w}_i \mathrm{~d} W,\quad\forall\;\; i\in\{0,\dots,N-1\}.
$$
To optimize the number of samples needed for a reliable approximation of the optimal control input, importance sampling is applied. This is done by multiplying the above equation with PDFs that are strictly positive when $q^\star(W) \boldsymbol{w}_i \neq 0$, resulting in:
\begin{align}
\bu_i^\star=\int \frac{q^\star(W)}{p(W)} \frac{p(W)}{q(W)} q(W) \boldsymbol{w}_i \mathrm{~d} W.
\label{eqn:opt_sto_int}
\end{align}
A weighting function, $\mu(W)$, is then defined as:
$$
\mu(W)=\frac{q^\star(W)}{p(W)} \frac{p(W)}{q(W)}.
$$
Incorporating this weighting function yields the optimal control input in \eqref{eqn:opt_sto_int} as
$
\bu_i^\star=\mathbb{E}_{\mathbb{Q}}\left[\mu(W) \boldsymbol{w}_i\right].
$
The ratios used in defining $\mu(W)$ are known and given by:
$$
\begin{aligned}
\frac{q^\star(W)}{p(W)} & =\frac{1}{\eta} \exp \left(-\frac{1}{\lambda} J(W)\right), \quad\;
\frac{p(W)}{q(W)} & =\exp \left(\sum_{i=0}^{N-1} \frac{1}{2} \bu_i^\mathrm{T} \Sigma^{-1}_\bu \bu_i-\boldsymbol{w}_i^\mathrm{T} \Sigma^{-1}_\bu \boldsymbol{w}_i\right).
\end{aligned}
$$
Substituting these into the weighting function equation results in:
$$
\begin{aligned}
q(W) & =\sum_{i=0}^{N-1} \frac{1}{2} \bu_i^T \Sigma^{-1}_\bu \bu_i-\boldsymbol{w}_i^{ T} \Sigma^{-1}_\bu \boldsymbol{w}_i, \quad
\mu(W) & =\frac{1}{\eta} \exp \left(-\frac{1}{\lambda} J(W)+q(W)\right).
\end{aligned}
$$
The normalizing constant $\eta$ is computationally challenging to determine directly. Instead, a Monte Carlo approach is used, employing $K$ realizations of control disturbances as follows:
$$
\eta \approx \sum_{k=1}^K \exp \left(-\frac{1}{\lambda} J\left(W^k\right)+q\left(W^k\right)\right).
$$
where $W^k$, $\mathcal{E}^k$ and $U^k$  is the $k^{\text{th}}$ sample given as follows:
$$
\begin{aligned}
\mathcal{E}^k & =\left\{\boldsymbol{\epsilon}_0^k, \cdots, \boldsymbol{\epsilon}_{N-1}^k\right\} ,\quad
U^k & =\left\{\bu_0^k, \cdots, \bu_{N-1}^k\right\},\quad
W^k & =U^k+\mathcal{E}^k.
\end{aligned}
$$
}
The MPPI algorithm then computes the importance weights for each trajectory based on their costs where the weight computed as $\mu_k$ is given by
\begin{align}
\mu_k  =\frac{1}{\eta} \exp \left(-\frac{1}{\lambda} J\left(W^k\right)+q(W^k)\right). \label{eqn:weights}
\end{align}
These weights give higher importance to trajectories with lower costs, effectively focusing the optimization on more promising regions of the control space.
Finally, the optimal control sequence is then computed as a weighted average of the sampled controls as $\bu_i^\star  ={\bu}_i+\sum_{k=1}^K \mu_k \epsilon_i^k$. In addition, the obtained control sequence is then smoothed using the Savitzky-Golay filter \cite{williams2018information}.
This formulation allows MPPI to approximate the optimal control by emphasizing low-cost trajectories while still maintaining exploration through the stochastic sampling process. 
\section{Proposed approach\label{sec:proposed_approach}}

In this section, we propose using an transformer architecture for initializing the MPPI with an informed mean control sequence followed by MPPI control. The overall approach involves two main phases: the learning phase and the execution phase. In the learning phase, the transformer model is trained using the teacher-forcing strategy on a data set that contains optimal control sequences from previous control tasks obtained using the original MPPI algorithm \cite{williams2018information}. The transformer architecture is specifically designed to capture long-horizon dependencies in control sequences, allowing it to effectively map the current state and environmental context to an initial mean control sequence. This sequence serves as an approximation that is closer to the optimal solution, thus improving the subsequent MPPI sampling process. In the execution phase, the trained transformer model is used for real-time prediction of the initial mean control sequence for the MPPI algorithm, using an autoregressive decoding approach to generate predictions. As shown in the results section (Section \ref{sec:results}), this informed mean initialization reduces the number of samples required for the \ours algorithm to converge to an optimal control sequence, thus accelerating convergence and improving computational performance.
\begin{figure}
    \centering
    \hfill
    \begin{minipage}{0.99\textwidth}
        \centering
        \scalebox{1.2}{
        \input{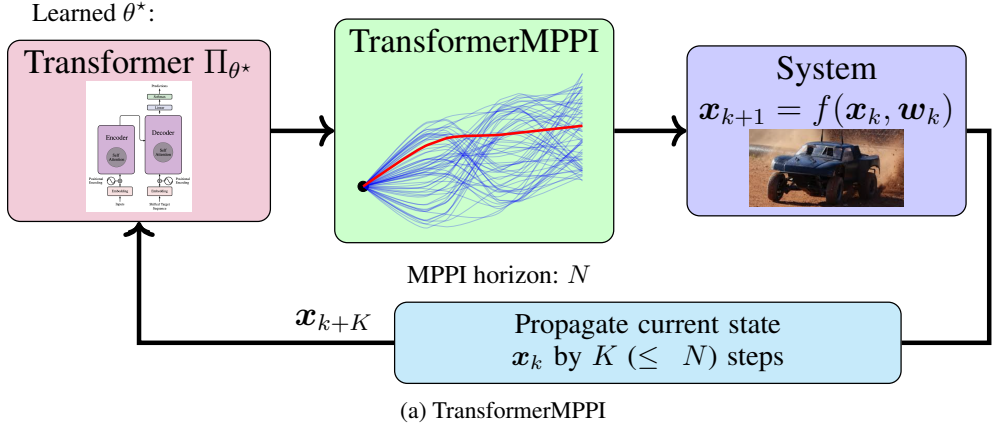}
        }
        \subcaption{\ours}
        \label{fig:}
    \end{minipage}
    \caption{\small  Schematic of \ours: Our approach enhances the computational efficiency of the MPPI framework by leveraging a transformer-based attention mechanism to generate informed initial mean control sequences, improving sample efficiency and accelerating convergence. This approach can be seamlessly integrated with any state-of-the-art MPPI algorithm. }
    \label{fig:enter-label}
\end{figure}
\subsection{Transformer architecture for informed mean control sequence initialization}
\label{subsec:transformer_control_prediction}
Let, the environment provides contextual information $\mathbf{c} \in \mathbb{R}^p$, such as obstacle coordinates for navigation 2D environment or lane information for autonomous racing and other parameters that are together stacked into a vector $\mathbf{c}$ with dimension $p$. Our goal is to learn a mapping from a sequence of past states and environmental context to a sequence of future control inputs given by:
\begin{equation*}
    \Pi_\theta: \left\{ \bx_{t-k}, \dots, \bx_{t}, \mathbf{c} \right\} \mapsto \left\{ \bu_{t}, \bu_{t+1}, \dots, \bu_{t+H-1} \right\},
\end{equation*}
where $k$ is the number of past time steps considered, $H$ is the prediction horizon, and $\theta$ are the parameters of the transformer. Towards this goal, we use a transformer architecture which use self-attention mechanisms to process input sequences and generate outputs. The proposed transformer consists of an encoder (Fig. \ref{fig:encoder}) and a decoder (Fig. \ref{fig:decoder}), each comprising multiple layers of self-attention and position-wise feedforward networks (FFN). The overall architecture is illustrated in Fig. \ref{fig:transformer}. The input to the encoder is a sequence of past states concatenated with the environmental context:
\begin{equation*}
    \bx_{\text{enc}} = \left[ \bx_{t-k+1}^\mathrm{T}, \dots, \bx_{t}^\mathrm{T}, \mathbf{c}^\mathrm{T} \right]^\mathrm{T} \in \mathbb{R}^{(k+1) \times n+p},
\end{equation*}
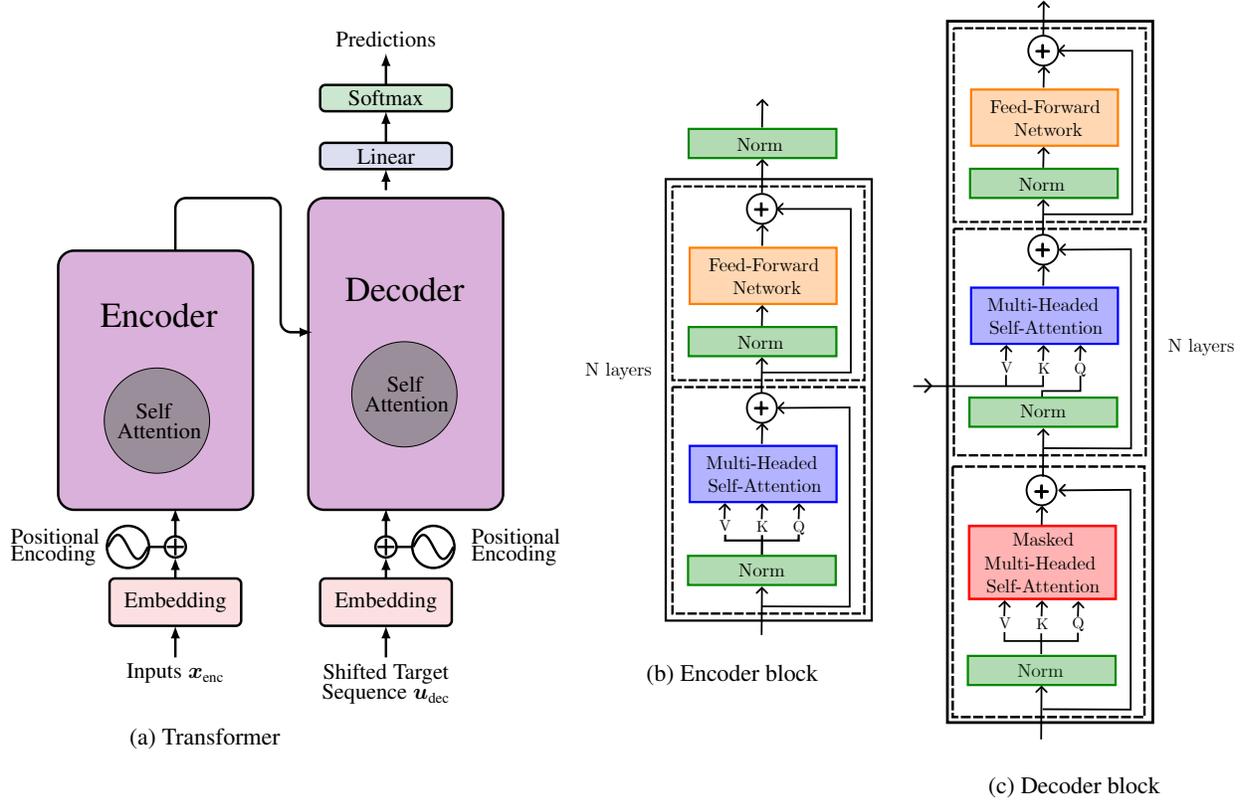
\begin{figure}
    \centering
        \begin{minipage}{0.33\textwidth}
        \centering
        \scalebox{0.7}{
        \begin{tikzpicture}
    \definecolor{emb_color}{RGB}{252,224,225}
    \definecolor{multi_head_attention_color}{RGB}{252,226,187}
    \definecolor{add_norm_color}{RGB}{242,243,193}
    \definecolor{ff_color}{RGB}{194,232,247}
    \definecolor{softmax_color}{RGB}{203,231,207}
    \definecolor{linear_color}{RGB}{220,223,240}
    \definecolor{gray_bbox_color}{RGB}{243,243,244}
    \definecolor{light_purple_color}{RGB}{218,177,218}

    \draw[fill=light_purple_color, line width=0.046875cm, rounded corners=0.300000cm] 
    (-0.975000, 6.230000) -- (2.725000, 6.230000) -- (2.725000, 1.305000) -- (-0.975000, 1.305000) -- cycle;
    \draw[fill=light_purple_color, line width=0.046875cm, rounded corners=0.300000cm] 
    (3.775000, 7.230000) -- (7.475000, 7.230000) -- (7.475000, 1.305000) -- (3.775000, 1.305000) -- cycle;

    \draw[line width=0.046875cm, fill=emb_color, rounded corners=0.100000cm] 
    (0.000000, 0.000000) -- (2.500000, 0.000000) -- (2.500000, -0.900000) -- (0.000000, -0.900000) -- cycle;
    \node[text width=2.500000cm, align=center] at (1.250000,-0.450000) {\large Embedding};

    \draw[line width=0.046875cm, fill=emb_color, rounded corners=0.100000cm] 
    (4.000000, 0.000000) -- (6.500000, 0.000000) -- (6.500000, -0.900000) -- (4.000000, -0.900000) -- cycle;
    \node[text width=2.500000cm, align=center] at (5.250000,-0.450000) {\large Embedding};

    \draw[line width=0.046875cm, fill=linear_color, rounded corners=0.100000cm] 
    (4.000000, 8.280000) -- (6.500000, 8.280000) -- (6.500000, 7.780000) -- (4.000000, 7.780000) -- cycle;
    \node[text width=2.500000cm, align=center] at (5.250000,8.030000) {\large Linear};

    \draw[line width=0.046875cm, fill=softmax_color, rounded corners=0.100000cm] 
    (4.000000, 9.380000) -- (6.500000, 9.380000) -- (6.500000, 8.880000) -- (4.000000, 8.880000) -- cycle;
    \node[text width=2.500000cm, align=center] at (5.250000,9.130000) {\large Softmax};

    \draw[line width=0.046875cm] (1.250000, 0.600000) circle (0.200000);
    \draw[line width=0.046875cm] (1.410000, 0.600000) -- (1.090000, 0.600000);
    \draw[line width=0.046875cm] (1.250000, 0.760000) -- (1.250000, 0.440000);

    \draw[line width=0.046875cm] (5.250000, 0.600000) circle (0.200000);
    \draw[line width=0.046875cm] (5.410000, 0.600000) -- (5.090000, 0.600000);
    \draw[line width=0.046875cm] (5.250000, 0.760000) -- (5.250000, 0.440000);

    \draw[line width=0.046875cm] (0.350000, 0.600000) circle (0.400000);

    \draw[line width=0.046875cm] (6.150000, 0.600000) circle (0.400000);

    \draw[-latex, line width=0.046875cm, rounded corners=0.2cm] 
        (1.25, 6.23) -- (1.25, 7.23) -- (3.25, 7.23) -- (3.25, 4.68) -- (3.85, 4.68);

    \draw[line width=0.046875cm, -latex] (5.250000, 7.380000) -- (5.250000, 7.780000);
    \draw[line width=0.046875cm, -latex] (5.250000, 8.280000) -- (5.250000, 8.880000);
    \draw[line width=0.046875cm, -latex] (1.250000, 0.000000) -- (1.250000, 0.400000);
    \draw[line width=0.046875cm, -latex] (1.250000, 0.800000) -- (1.250000, 1.330000);
    \draw[line width=0.046875cm, -latex] (5.250000, 0.800000) -- (5.250000, 1.330000);
    \draw[line width=0.046875cm, -latex] (5.250000, 0.000000) -- (5.250000, 0.400000);

    \draw[line width=0.046875cm] (0.750000, 0.600000) -- (1.050000, 0.600000);
    \draw[line width=0.046875cm] (5.450000, 0.600000) -- (5.750000, 0.600000);
    \draw[line width=0.046875cm] 
        (-0.030000, 0.600000) 
        -- (-0.014490, 0.629156) 
        -- (0.001020, 0.657833) 
        -- (0.016531, 0.685561) 
        -- (0.032041, 0.711884) 
        -- (0.047551, 0.736369) 
        -- (0.063061, 0.758616) 
        -- (0.078571, 0.778258) 
        -- (0.094082, 0.794973) 
        -- (0.109592, 0.808486) 
        -- (0.125102, 0.818576) 
        -- (0.140612, 0.825077) 
        -- (0.156122, 0.827883) 
        -- (0.171633, 0.826946) 
        -- (0.187143, 0.822284) 
        -- (0.202653, 0.813971) 
        -- (0.218163, 0.802145) 
        -- (0.233673, 0.786999) 
        -- (0.249184, 0.768783) 
        -- (0.264694, 0.747796) 
        -- (0.280204, 0.724382) 
        -- (0.295714, 0.698925) 
        -- (0.311224, 0.671845) 
        -- (0.326735, 0.643584) 
        -- (0.342245, 0.614608) 
        -- (0.357755, 0.585392) 
        -- (0.373265, 0.556416) 
        -- (0.388776, 0.528155) 
        -- (0.404286, 0.501075) 
        -- (0.419796, 0.475618) 
        -- (0.435306, 0.452204) 
        -- (0.450816, 0.431217) 
        -- (0.466327, 0.413001) 
        -- (0.481837, 0.397855) 
        -- (0.497347, 0.386029) 
        -- (0.512857, 0.377716) 
        -- (0.528367, 0.373054) 
        -- (0.543878, 0.372117) 
        -- (0.559388, 0.374923) 
        -- (0.574898, 0.381424) 
        -- (0.590408, 0.391514) 
        -- (0.605918, 0.405027) 
        -- (0.621429, 0.421742) 
        -- (0.636939, 0.441384) 
        -- (0.652449, 0.463631) 
        -- (0.667959, 0.488116) 
        -- (0.683469, 0.514439) 
        -- (0.698980, 0.542167) 
        -- (0.714490, 0.570844) 
        -- (0.730000, 0.600000);
        \draw[line width=0.046875cm] 
        (5.770000, 0.600000) 
        -- (5.785510, 0.629156) 
        -- (5.801020, 0.657833) 
        -- (5.816531, 0.685561) 
        -- (5.832041, 0.711884) 
        -- (5.847551, 0.736369) 
        -- (5.863061, 0.758616) 
        -- (5.878571, 0.778258) 
        -- (5.894082, 0.794973) 
        -- (5.909592, 0.808486) 
        -- (5.925102, 0.818576) 
        -- (5.940612, 0.825077) 
        -- (5.956122, 0.827883) 
        -- (5.971633, 0.826946) 
        -- (5.987143, 0.822284) 
        -- (6.002653, 0.813971) 
        -- (6.018163, 0.802145) 
        -- (6.033673, 0.786999) 
        -- (6.049184, 0.768783) 
        -- (6.064694, 0.747796) 
        -- (6.080204, 0.724382) 
        -- (6.095714, 0.698925) 
        -- (6.111224, 0.671845) 
        -- (6.126735, 0.643584) 
        -- (6.142245, 0.614608) 
        -- (6.157755, 0.585392) 
        -- (6.173265, 0.556416) 
        -- (6.188776, 0.528155) 
        -- (6.204286, 0.501075) 
        -- (6.219796, 0.475618) 
        -- (6.235306, 0.452204) 
        -- (6.250816, 0.431217) 
        -- (6.266327, 0.413001) 
        -- (6.281837, 0.397855) 
        -- (6.297347, 0.386029) 
        -- (6.312857, 0.377716) 
        -- (6.328367, 0.373054) 
        -- (6.343878, 0.372117) 
        -- (6.359388, 0.374923) 
        -- (6.374898, 0.381424) 
        -- (6.390408, 0.391514) 
        -- (6.405918, 0.405027) 
        -- (6.421429, 0.421742) 
        -- (6.436939, 0.441384) 
        -- (6.452449, 0.463631) 
        -- (6.467959, 0.488116) 
        -- (6.483469, 0.514439) 
        -- (6.498980, 0.542167) 
        -- (6.514490, 0.570844) 
        -- (6.530000, 0.600000);

    \draw[line width=0.046875cm, -latex] (1.250000, -1.500000) -- (1.250000, -0.900000);
    \draw[line width=0.046875cm, -latex] (5.250000, -1.500000) -- (5.250000, -0.900000);
    \draw[line width=0.046875cm, -latex] (5.250000, 9.380000) -- (5.250000, 9.980000);

    \node[text width=2.500000cm, anchor=north, align=center] at (1.250000,-1.500000) {\large Inputs $\bx_{\text{enc}}$};
    \node[text width=2.500000cm, anchor=north, align=center] at (5.250000,-1.500000) {\large Shifted Target Sequence $\bu_{\text{dec}}$};
    \node[text width=2.500000cm, anchor=south, align=center] at (5.250000,9.980000) {\large Predictions};

    \node[text width=2.000000cm, anchor=east] at (0.250000,0.600000) {\large Positional \vspace{-0.05cm} \linebreak Encoding};
    \node[text width=2.000000cm, anchor=west] at (6.750000,0.600000) {\large Positional \vspace{-0.05cm} \linebreak Encoding};

    \node[text width=3.000000cm, anchor=west] at (-0.30000,5.00000) {\fontsize{19pt}{21pt}\selectfont Encoder};
    \node[text width=3.000000cm, anchor=west] at (4.350000,5.50000) {\fontsize{19pt}{21pt}\selectfont Decoder};

        \draw[fill=gray, fill opacity=0.7] (0.9, 3) circle [radius=1];
    \node[text width=1.3cm] at (0.8, 3) {\fontsize{12pt}{16pt}\selectfont \hspace{0.35cm}Self \\ Attention};

  \draw[fill=gray, fill opacity=0.7] (5.6, 3.5) circle [radius=1];
    \node[text width=1.3cm] at (5.5, 3.5) {\fontsize{12pt}{16pt}\selectfont \hspace{0.3cm} Self\\ Attention};
\end{tikzpicture}
        }
        \subcaption{Transformer}
        \label{fig:transformer}
    \end{minipage}
    \hfill
    \begin{minipage}{0.27\textwidth}
        \centering

        \includegraphics[width = \linewidth]{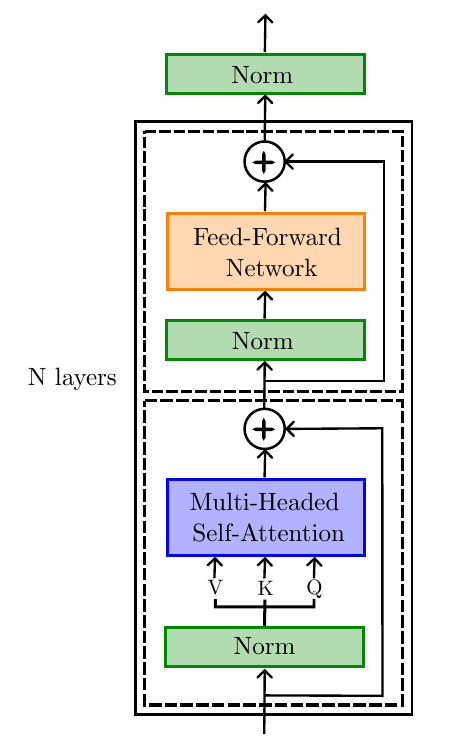}
        \subcaption{Encoder block}
        \label{fig:encoder}
    \end{minipage}
      \begin{minipage}{0.27\textwidth}
        \centering
         \includegraphics[width = \linewidth]{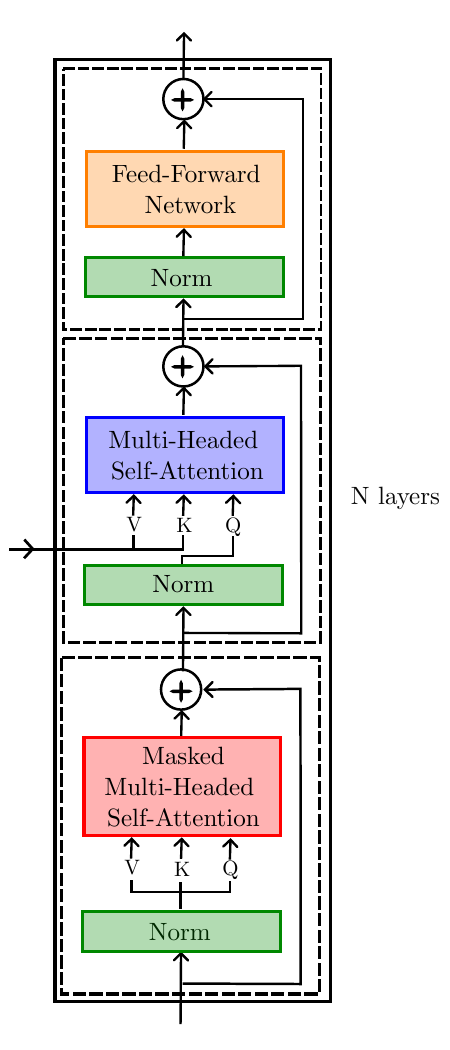}
        \subcaption{Decoder block}
        \label{fig:decoder}
    \end{minipage}
    \caption{\small Transformer architecture for mean control sequence prediction}
    \label{fig:transformer_prediction}
\end{figure}
where each state $\bx_i$ and the context $\mathbf{c}$ are vectors. We augment the sequence with the context $\mathbf{c}$ to provide the model with static environmental information throughout the prediction. During training, the decoder input is a sequence of ground truth shifted control input sequence:
\begin{equation*}
    \bu_{\text{dec}} = \left[ \textbf{0}, \bu_{t}^\mathrm{T}, \bu_{t+1}^\mathrm{T}, \dots, \bu_{t+H-2}^\mathrm{T} \right]^\mathrm{T} \in \mathbb{R}^{H \times m},
\end{equation*}
where the first value (i.e., start token) is initialized to $\textbf{0}$  and the last value of control input sequence ignored. Assuming $\text{Embed}_{\text{enc}}$ and $\text{Embed}_{\text{dec}}$ are linear transformations, we pass the encoder and decoder inputs through embedding layers to lift them into a higher-dimensional space as follows:
\begin{align*}
    \mathbf{E}_{\text{enc}} = \text{Embed}_{\text{enc}}(\bx_{\text{enc}}) \in \mathbb{R}^{(k+1) \times \mathrm{d}_{\text{model}}}, \quad \mathbf{E}_{\text{dec}} = \text{Embed}_{\text{dec}}(\bu_{\text{dec}}) \in \mathbb{R}^{H \times \mathrm{d}_{\text{model}}},
\end{align*}
where $\mathrm{d}_{\text{model}}$ indicates the dimensionality of the model. To incorporate sequence order information, we add positional encodings to the embedded inputs. If $\text{pos}$ represents the position in the sequence (ranging from $0$ to $L-1$), the positional encoding $\mathbf{P} \in \mathbb{R}^{L \times \mathrm{d}_{\text{model}}}$ for dimension $i$ is defined as:
\begin{align*}
    \mathbf{P}_{\mathrm{pos, 2i}} = \sin\left( \frac{\mathrm{pos}}{10000^{\frac{\mathrm{2i}}{\mathrm{d}_{\text{model}}}}} \right), \quad\mathbf{P}_{\mathrm{pos, 2i+1}} = \cos\left( \frac{\mathrm{pos}}{10000^{\frac{2i}{\mathrm{d}_{\text{model}}}}} \right),
\end{align*}
where $i$ is the dimension index (ranging from $0$ to $\mathrm{d}_{\text{model}}/2 - 1$). The positional encodings are then added to the embedded inputs as follows:
\begin{align*}
    \mathbf{Z}_{\text{enc}} = \mathbf{E}_{\text{enc}} + \mathbf{P}_{\text{enc}}, \quad \mathbf{Z}_{\text{dec}} = \mathbf{E}_{\text{dec}} + \mathbf{P}_{\text{dec}}.
\end{align*}
Both the encoder and decoder are composed of $N$ identical layers. Each layer in encoder includes a multi-head self-attention framework and a position-wise $\texttt{FFN}$. On the other hand, each decoder layer contains a masked multi-head self-attention framework, an encoder-decoder attention framework, and a position-wise $\texttt{FFN}$. Note that residual connections and layer normalization are implemented following each sub-layer. The multi-head attention framework allows the model to focus on information from various representation subspaces. Let queries $\mathbf{Q}$, keys $\mathbf{K}$, and values $\mathbf{V}$ be given, then the scaled dot-product attention can be computed as follows:
\begin{equation*}
    \texttt{Attention}(\mathbf{Q}, \mathbf{K}, \mathbf{V}) = \texttt{softmax}\left( \frac{\mathbf{Q} \mathbf{K}^\mathrm{T}}{\sqrt{d_k}} \right) \mathbf{V},
\end{equation*}
where $d_k$ is the dimensionality of the keys. For multi-head attention with $h$ heads, queries, keys, and values are each linearly transformed $h$ times using distinct learned linear projections into corresponding spaces of dimensions $d_k$, $d_k$, and $d_v$, respectively. 
The outputs are concatenated and projected again to obtain the final values:
\begin{align*}
    \texttt{MultiHead}(\mathbf{Q}, \mathbf{K}, \mathbf{V}) &= \texttt{Concat}(\text{head}_1, \dots, \text{head}_h) \mathbf{W}^O, 
\end{align*}
$\text{where } \text{head}_i = \text{Attention}(\mathbf{Q} \mathbf{W}_i^Q, \mathbf{K} \mathbf{W}_i^K, \mathbf{V} \mathbf{W}_i^V)$ with learnable projection matrices $\mathbf{W}_i^Q$, $\mathbf{W}_i^K$, $\mathbf{W}_i^V$, and $\mathbf{W}^O$. Each layer includes a position-wise $\texttt{FFN}$ that is applied independently and uniformly to each position, as described below:
\begin{equation*}
    \texttt{FFN}(\bx) = \texttt{ReLU}(\bx \mathbf{W}_1 + \mathbf{b}_1) \mathbf{W}_2 + \mathbf{b}_2,
\end{equation*}
where $\mathbf{W}_1$, $\mathbf{W}_2$, $\mathbf{b}_1$, and $\mathbf{b}_2$ are learnable parameters. The decoder outputs are then passed through a final linear layer to generate the predicted control sequences:
\begin{equation*}
    \hat{\bu} = \texttt{Linear}(\mathbf{Z}_{\text{dec}}) \in \mathbb{R}^{H \times m}.
\end{equation*}
\subsection{\ours Algorithm} \label{subsec
} 
 In standard MPPI control, the initialization of the mean control sequence, denoted as \( \bu_t \), is frequently set to zero or based on the previous control sequence. Although this approach may suffice in relatively static or simplistic scenarios, it often proves suboptimal in dynamic or high-dimensional environments.
In this paper, transformers are chosen over recurrent neural networks such as gated recurrent units \cite{cho2014learning, zinage2024leveraging} due to several key advantages. Firstly, transformers excel at capturing long-range dependencies within sequences through their self-attention mechanisms, which dynamically weigh the importance of different elements in the input sequence. This capability is crucial in dynamic and high-dimensional environments where understanding intricate temporal and contextual relationships can significantly improve performance. Secondly, transformers support parallel processing of sequence data, leading to faster training and inference times compared to the sequential nature of RNNs. This parallelization is particularly beneficial for real-time applications where computational efficiency is paramount. These attributes make transformers a more suitable and effective framework for generating informed control sequences in the context of MPPI control.
Building on these strengths, \ours uses a transformer-based model to generate an informed initialization of the mean control sequence, leveraging contextual information about the current state and environment.
Specifically, at each time step $t$, the transformer predicts the sequence of mean control actions for the planning horizon $H$ as follows: 
$
\{\hat{\bu}_{t
},\;\hat{\bu}_{t
+1},\dots,\hat{\bu}_{t
+H}\} = \Pi_\theta\left( \left\{ \bx_{t-k}, \dots, \bx_{t}, \mathbf{c} \right\} \right)$.
where $\Pi_\theta$ represents the trained transformer model parameterized by $\theta$. Subsequently, control samples for the MPPI algorithm are drawn around the transformer-predicted mean sequence as:
$ \bu_{t
+i}^k = \hat{\bu}_{t
+i} + \boldsymbol{\epsilon}_{t
+i}^k$ where $\boldsymbol{\epsilon}_{t
+i}^k \sim \mathcal{N}(0, \Sigma_\bu)$ is the control noise for sample $k$, $\Sigma_\bu$ is the covariance matrix, and $i\in\{0,\dots,H\}$.
Incorporating transformer-based initialization introduces a significant improvement in the efficiency of the standard MPPI algorithm \cite{williams2018information}. By centering the sampling process around control sequences that are more likely to result in lower cost outcomes, the search space is effectively constrained to regions with higher probability of success. This not only improves sampling efficiency, but also reduces the computational burden associated with evaluating suboptimal control sequences. A comprehensive description of the implementation steps, including the integration of transformer predictions, is provided in Algorithm~\ref{alg:ours}.

\begin{algorithm}[htbp] \caption{\ours} \label{alg} 

\begin{algorithmic}[1] 
\Require Transition model $f$ \eqref{eqn:nonlinear_dynamics}, cost functions $q$, $\phi$, number of samples $K$, time horizon $H$, transformer model $\Pi_\theta$ (parameters $\theta$), initial state $\bx_0$, context $\mathbf{c}$ 
\State Initialize control sequence $\{ \bu_0, \dots, \bu_{H-1} \} \leftarrow \mathbf{0}$ 
\State Initialize state $\bx \leftarrow \bx_0$ \While{goal not reached} 
\State Obtain current state $\bx_t$ 
\State Predict mean control sequence using transformer: $\{\hat{\bu}_{t+i}\}_{i=0}^{H} \leftarrow \Pi_\theta\left( \left\{ \bx_{t-k}, \dots, \bx_{t}, \mathbf{c} \right\} \right) $
\For{$k = 1$ to $K$} 
\State Sample control noise $\boldsymbol{\epsilon}_{t
+i}^k \sim \mathcal{N}(0, \Sigma_\bu)$ where $i\in\{0,\dots,H\}$
\State Generate control sequence: $ \bu_{t
+i}^k = \hat{\bu}_{t
+i} + \boldsymbol{\epsilon}_{t
+i}^k\nonumber $ where $i\in\{0,\dots,H\}$
\State Simulate trajectory using $f$ i.e., 
$ \bx_{t+1}^k = f\left( \bx_t, \bu_{t
+H}^k \right)\nonumber 
 $
\State Compute cost $ J_k $ using \eqref{eqn:objective_function} \EndFor 
\State Compute weights $ \mu_k$ using \eqref{eqn:weights}
\State Update control sequence: $ \bu_{t+i} \leftarrow \hat{\bu}_{t+i} + \sum_{k=1}^{K} \mu_k \boldsymbol{\epsilon}_{t
+i}^k$ where $i\in\{0,\dots,H\}$
\State Apply control $\bu_t$ to \eqref{eqn:nonlinear_dynamics} 
\State $t \leftarrow t + 1$ 
\EndWhile 
\end{algorithmic} 
\label{alg:ours}
\end{algorithm}

\subsection{Data collection and training}

To effectively train the transformer, a diverse dataset is collected to include the wide array of scenarios a system may encounter during deployment. Specifically, $N_\text{env}$ distinct environments were generated, each characterized by a random arrangement of parameterized obstacles. 
The positions of these obstacles were sampled from predefined probability distributions to ensure significant variability across scenarios.
For each environment, optimal trajectories were generated using the MPPI controller \cite{williams2018information}. These trajectories formed the basis for the training dataset, with recorded sequences comprising states, control inputs, and environmental context structured as follows:
\begin{equation}
    \left\{ \left( \bx_t^i, \bu_t^i, \mathbf{c}^i \right) \right\}_{t=0}^{T_i}, \quad i = \{1, \dots, N_\text{env}\},
\end{equation}
where $T_i $ denotes the length of the $i$-th trajectory. From each trajectory, input-output pairs were extracted for training purposes. The input sequence consisted of the past $k$ states and the corresponding environmental context, while the output was defined as the future control sequence of length $H$.
Before training, we normalize all datasets using the quantile transform method. This normalization converts the data into a uniform distribution, which is especially beneficial when the original distribution is unknown or does not conform to the Gaussian distribution often assumed by many machine learning algorithms. Additionally, this normalization technique improves the robustness to outliers by ranking data points instead of directly scaling their values. We set the loss function as Huber loss, and the optimizer used is Adam with hyperparameters tuned for efficient convergence. We use early stopping of loss with patience of 50 epochs to prevent the model from overfitting. By training on a diverse set of scenarios, the transformer learns to generalize mapping from past states and environmental contexts to future control sequences. This capability allows for real-time deployment in robotic systems as forward passes through the trained transformer are computationally efficient. A summary of the configurations used for data generation and transformer training is provided in Table \ref{tab:configuration}.

\begin{table}[htbp]
    \centering
     \caption{Configurations used for data generation and training}
           \label{tab:configuration}
    \begin{tabular}{|c|c|c|}
    \hline
        \textbf{Configuration} & \textbf{Navigation 2D} & \textbf{Autonomous Racing} \\
    \hline
         Episodes used for training $N_{\text{env}}$ (max steps $T_i$) & 1000 (150) & 300 (500) \\
    \hline
         Number of obstacles (radius (in m)) & 15 (1) & 50 (0.8) \\
    \hline
         Horizon $H$ & 20 & 25 \\
    \hline
         $k$ & 5 & 5 \\
    \hline
        \makecell{Constraints on control inputs} & \makecell{$\bu_1 = [0, 2] \text{m/s}$ \\ $\bu_2 = [-1, 1] \text{rad/s}$}& \makecell{$\bu_1 = [-2, 2] \text{m/$s^2$}$ \\ $\bu_2 = [-0.25, 0.25] \text{rad}$}\\
    \hline
         Batch size & 256 & 256 \\
    \hline
         Learning rate & 0.0005 & 0.0005 \\
    \hline
         Number of epochs (patience) & 2000 (50) & 2000 (50) \\
    \hline
         \makecell{Transformer parameters \\ (hidden size, encoder/decoder layers, heads, dropout)} & 
         \makecell{(256, 3, 8, 0.1)} & 
         \makecell{(256, 3, 8, 0.1)} \\
    \hline          
    \end{tabular}
\end{table}
Note that MPPI is a stochastic controller, which means that the optimal control sequence generated for a given configuration is not necessarily unique. This inherent variability arises due to the stochastic sampling of control inputs, which can lead to different sequences even under identical conditions. To address this issue within the context of \ours, which learns a deterministic mapping, we take the necessary precaution of setting random seeds at critical stages of the experiment. This ensures consistency and reproducibility in the comparison between \ours and MPPI. By controlling the randomness inherent in MPPI, the experiments can focus on evaluating the core differences in performance attributable to the algorithmic approach rather than the stochastic nature of the baseline.

\section{Results}
\label{sec:results}
We compare our proposed approach, \ours, with the MPPI \cite{williams2018information} in this section. Through our numerical experiments, we aim to answer the following questions: (i) what is the average reduction of the cost observed using our approach as the number of samples and the episodes are varied? (ii) what is the overall decrease in average steps taken by the agent to reach the goal/terminal state with our approach as the number of samples and the episodes are varied compared with the baseline method? (iii) finally, is our approach generalizable to a varying number of dynamic obstacles with respect to the average cost. We consider two environments where we test \ours, first is a Navigation 2D environment with obstacles and second is the problem of autonomous racing where, given the lane information, the task is to reach the terminal state while following the lane as closely as possible. All benchmarking experiments were performed using PyTorch library on a desktop equipped with an Intel(R) Core(TM) i9-10900K CPU @ 3.70GHz and an NVIDIA RTX A4000 GPU with 16 GB of GDDR6 memory. Source code will be made available at \href{https://github.com/shrenikvz/transformer-mppi}{https://github.com/shrenikvz/transformer-mppi}.

\subsection{Navigation 2D\label{subsec:2d_navigation}}
\label{subsec:results_navigation}

\begin{figure}[H]
    \centering
    \newcommand{\subfigwidth}{0.45\textwidth}
    \newcommand{\subfigheight}{6cm} 
    
    \begin{subfigure}[b]{\subfigwidth}
        \centering
        \resizebox{\textwidth}{!}{
            \begin{tikzpicture}
\begin{axis}[
    scale=0.8,
    xmin=1,
    xmax=20,
    ymin=1,
    ymax=2.5, 
    xlabel={\huge Horizon},
    ylabel={\huge Velocity (m/s)},
    grid=both,
    legend style={
    font=\huge,
        at={(0.5,-0.25)},
        anchor=north, legend columns=-1
    },
    x=0.8cm, 
    mark size=1.5pt 
]

\pgfplotstableread[col sep=comma]{csv/transformer_predictions_2d_obstacle_u1.csv}\datatable

\addplot[thick, color=blue, mark=*] 
    table[x=horizon, y= original] {\datatable};
\addplot[thick, color=red, mark=square*] 
    table[x=horizon, y=transformer] {\datatable};

\legend{MPPI, Predicted}
\end{axis}
\end{tikzpicture}
        }
        \caption{Velocity $\bu_1$ versus horizon}
    \end{subfigure}
    \begin{subfigure}[b]{\subfigwidth}
        \centering
        \resizebox{\textwidth}{!}{
            \begin{tikzpicture}
\begin{axis}[
    scale=0.8,
    xmin=1,
    xmax=20,
    ymin=-1.5,
    ymax=1.5, 
    xlabel={\huge Horizon},
    ylabel={\huge Angular rate (rad/s)},
    grid=both,
    legend style={
    font=\huge,
        at={(0.5,-0.25)},
        anchor=north, legend columns=-1
    },
    x=0.8cm, 
    mark size=1.5pt 
]

\pgfplotstableread[col sep=comma]{csv/transformer_predictions_2d_obstacle_u2.csv}\datatable

\addplot[thick, color=blue, mark=*] 
    table[x=horizon, y= original] {\datatable};
\addplot[thick, color=red, mark=square*] 
    table[x=horizon, y=transformer] {\datatable};

\legend{MPPI, Predicted}
\end{axis}
\end{tikzpicture}
        }
        \caption{Angular rate $\bu_2$ versus horizon}
    \end{subfigure}
    \caption{Transformer predictions for a random sample from test data for navigation 2D}
    \label{fig:transformer_predictions_2d_navigation}
\end{figure}
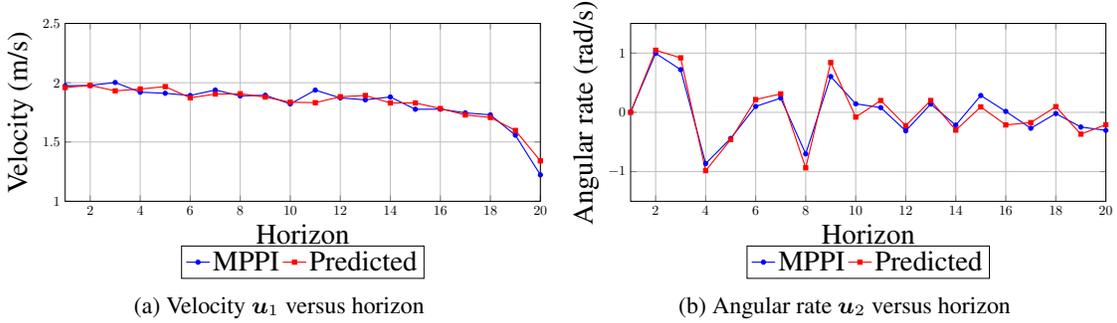
\begin{figure}[H]
    \centering
    \newcommand{\subfigwidth}{0.45\textwidth}
    \newcommand{\subfigheight}{6cm} 
    
    \begin{subfigure}[b]{\subfigwidth}
        \centering
        \resizebox{\textwidth}{!}{
        \includegraphics[width=0.9\linewidth]{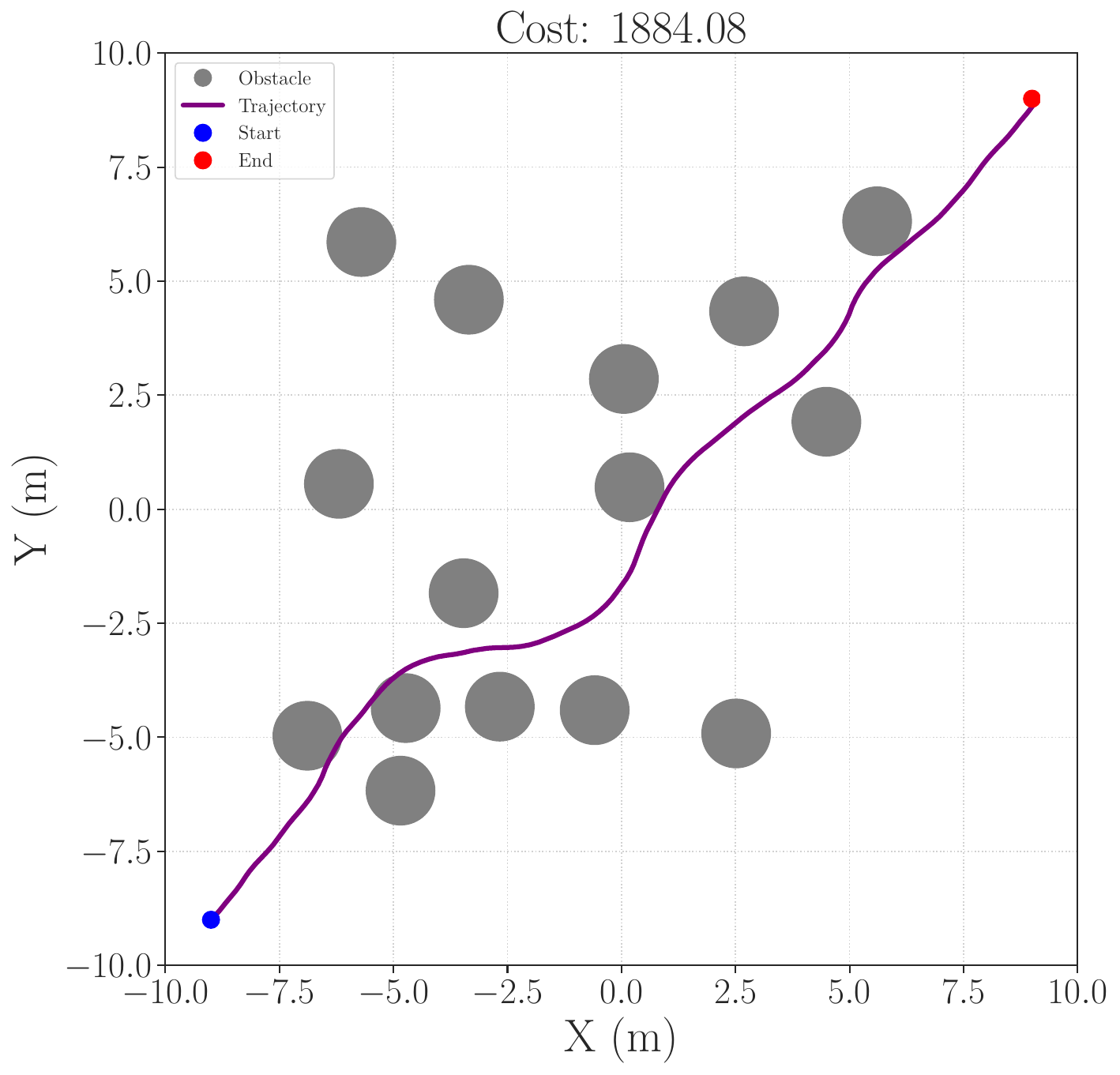}
        }
        \caption{MPPI}
    \end{subfigure}
    \begin{subfigure}[b]{\subfigwidth}
        \centering
        \resizebox{\textwidth}{!}{
            \includegraphics[width = 0.9\linewidth]{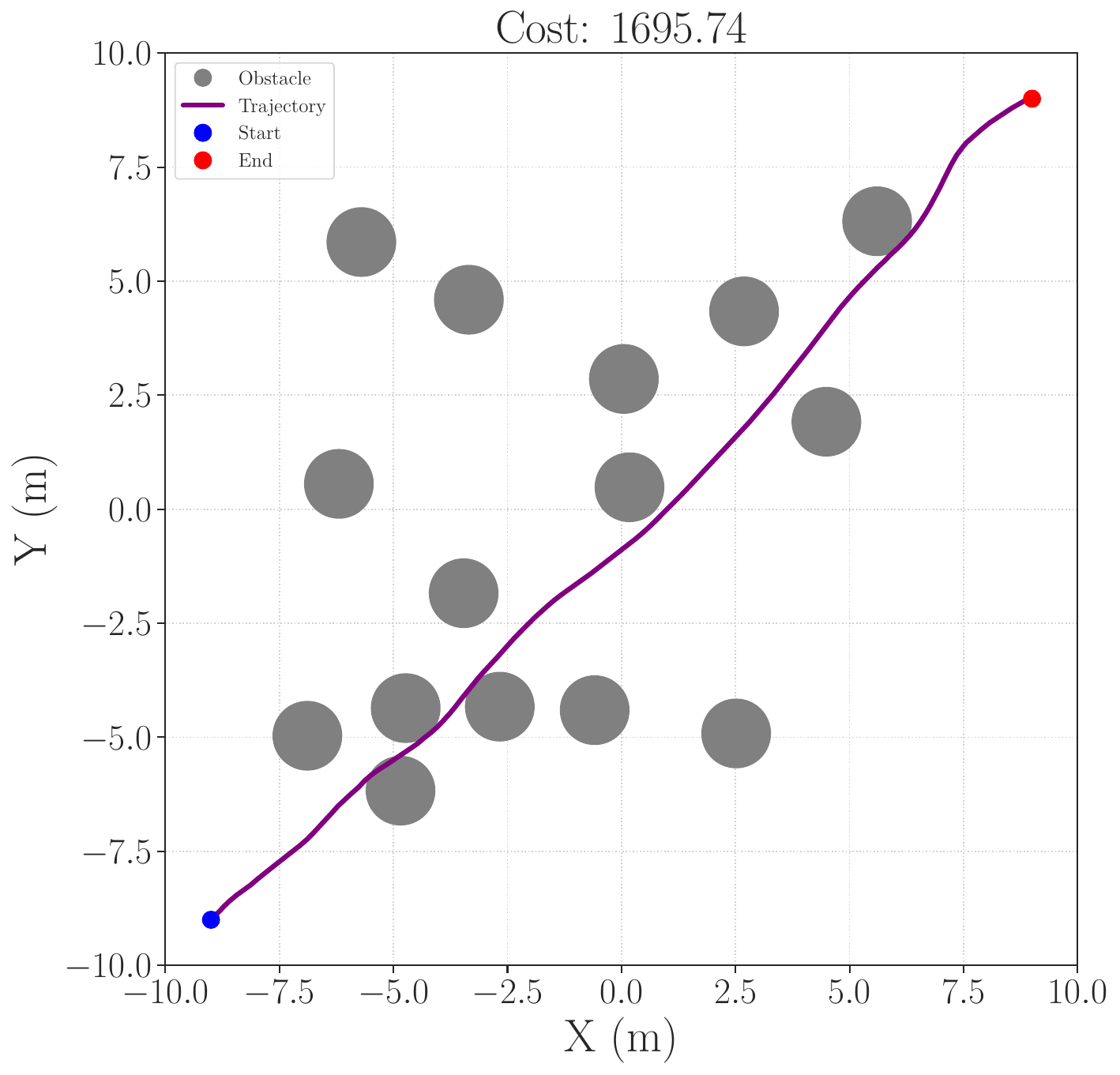}
        }
        \caption{\ours}
    \end{subfigure}
    
    \caption{Trajectories generated via MPPI and \ours for navigation 2D when the number of samples are low (i.e, 50). 
    }
    \label{fig:trajectory_2d_navigation_map}
\end{figure}

For the navigation in a 2D environment, we simulate a robotic agent through a 2D workspace consisting of static and dynamic obstacles. The goal is to reach a predefined terminal state while minimizing the cost associated with the trajectory. The agent’s control space is $\bu=[v,\omega]^\mathrm{T}$ where $v$ and $\omega$ are the velocity and angular rate respectively. The cost function was defined as $J(\bx_k) =\sum_{i=k}^{N+k-1}J_{\text{goal}}(\bx_i) + 10000 J_{\text{obstacle}}(\bx_i) $
where the goal cost
$J_{\text{\text{goal}}}(\bx_k) = \|\bx_k - \bx_{\text{\text{goal}}}\|_2$
where the state at time $k$ is $\bx_k=[x,y,\theta]^\mathrm{T}$,
$\bx_{\text{\text{goal}}}$ is the goal position and the obstacle cost $J_{\text{obstacle}}$ is equal to one if the agent collides with an obstacle or is out of bounds and zero otherwise.  Fig. \ref{fig:trajectory_2d_navigation_map} compares the trajectories taken by MPPI and \ours in the presence of 15 obstacles when the number of samples is 50. We can see that with fewer samples, the trajectory taken by \ours is much smoother and straight compared to that of standard MPPI. 
\subsubsection*{Cost analysis}
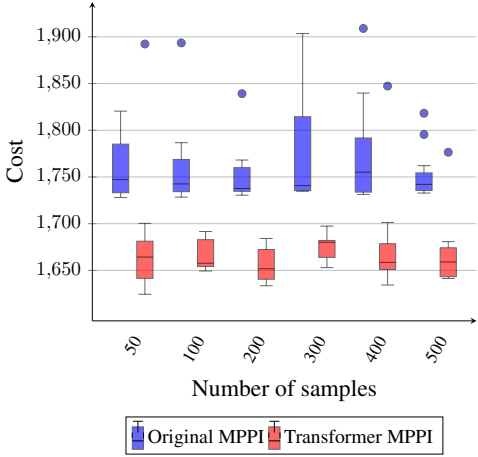
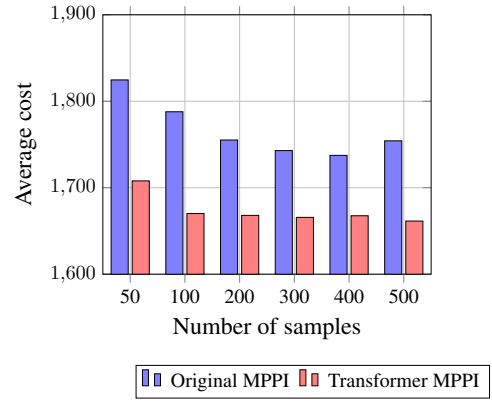
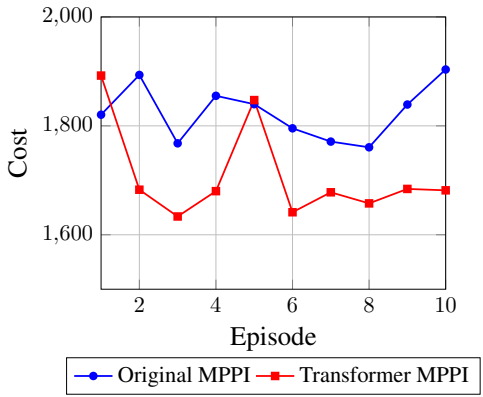
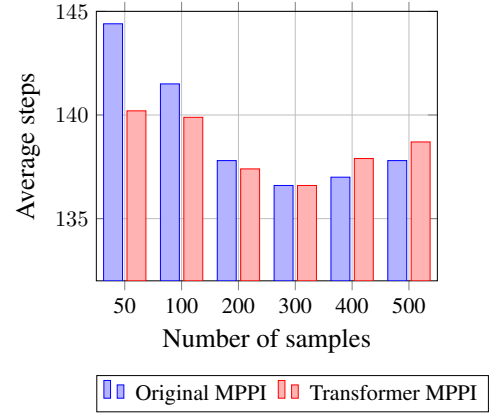
\begin{figure}[htbp]
  \centering
    \newcommand{\subfigwidth}{0.4\textwidth}
    \newcommand{\subfigheight}{4cm} 

    \begin{subfigure}[b]{\subfigwidth}
        \centering
        \resizebox{\textwidth}{!}{

\pgfplotsset{
    boxplot legend/.style={
        legend image code/.code={
            \draw[fill=#1, opacity = 0.6, draw=black] (0cm,-0.2cm) rectangle (0.2cm,0.2cm); 
            \draw[black] (0cm,0cm) -- (0cm,0.1cm); 
            \draw[black] (0.2cm,0cm) -- (0.2cm,0.1cm); 
            \draw[black] (0cm,-0.05cm) -- (0.2cm,-0.05cm); 
            \draw[black] (0.01cm,0.3cm) -- (0.19cm,0.3cm); 
            \draw[black] (0.1cm,0.1cm) -- (0.1cm,0.3cm); 
        }
    }
}
\begin{tikzpicture}
    \pgfplotstableread[col sep=comma]{data_mppi.csv}\csvdataone
    \pgfplotstableread[col sep=comma]{data_transformermppi.csv}\csvdatatwo

    \begin{axis}[
        boxplot/draw direction = y,
        axis x line=bottom, 
        axis y line=left,   
        enlarge y limits,
        ymajorgrids,
        xlabel = {\large Number of samples},
        xtick = {2,5,8,11,14,17}, 
        xticklabel style = {align=center, font=\small, rotate=60},
        xticklabels = {50, 100, 200, 300, 400, 500},
        xtick style = {draw=none}, 
        ylabel = {\large Cost},
        ytick = {1650, 1700, 1750, 1800, 1850, 1900},
        xmin=0, xmax=19,
        clip=false, 
        legend style={
            at={(0.5,-0.3)}, 
            anchor=north,
            legend columns=2, 
            every even column/.append style={column sep=0.4cm}, 
        },
        legend cell align={left}, 
    ]
        \pgfmathsetmacro{\shift}{0.6}
        
        \foreach \n in {0,...,5} { 
            \pgfmathtruncatemacro{\sample}{\n +1} 
            \pgfmathsetmacro{\xcenter}{(\sample-1)*3 +2} 
                
            \addplot+[
                boxplot,
                fill=blue,
                opacity=0.6,
                draw=black,
                boxplot/draw position={\xcenter - \shift},
                forget plot, 
            ] table[y index=\n] {\csvdataone};
            
            \addplot+[
                boxplot,
                fill=red,
                opacity = 0.6,
                draw=black,
                boxplot/draw position={\xcenter + \shift},
                forget plot, 
            ] table[y index=\n] {\csvdatatwo};
        }
        
        \addlegendimage{boxplot legend=blue}
        \addlegendentry{Original MPPI}

        \addlegendimage{boxplot legend=red}
        \addlegendentry{Transformer MPPI}
    \end{axis}
\end{tikzpicture}
        }
        \caption{Box plot of cost vs number of samples (over successful episodes).}
        \label{fig:box_plot_vs_number_of_samples_navigation}
    \end{subfigure}
    \hfill
    \begin{subfigure}[b]{\subfigwidth}
        \centering
        \resizebox{\textwidth}{!}{
            \begin{tikzpicture}
\begin{axis}[
    scale=0.8,
    ybar,
    ymin=1600,
    ymax=1900, 
    bar width=0.3cm, 
    legend style={
        at={(0.1,-0.35)},
        anchor=north west,
        legend columns=-1,
        column sep=0.3em 
    },
    ylabel={\large Average cost},
    xlabel={\large Number of samples},
    symbolic x coords={50, 100, 200, 300, 400, 500}, 
    xtick=data,
    x=1.2cm, 
    grid=both, 
]

\pgfplotstableread[col sep=comma]{csv/average_cost_vs_num_samples_obstacle.csv}\datatable

\addplot[
    fill=blue!50, 
] 
table[x=num_samples, y=original_mppi] {\datatable};

\addplot[
    fill=red!50, 
] 
table[x=num_samples, y=transformer_mppi] {\datatable};

\legend{Original MPPI, Transformer MPPI}
\end{axis}
\end{tikzpicture}
        }
        \caption{Average cost vs. number of samples (over successful episodes).}
        \label{fig:avg_cost_vs_number_of_samples_navigation}
    \end{subfigure}
        \begin{subfigure}[b]{\subfigwidth}
        \centering
        \resizebox{\textwidth}{!}{
            \begin{tikzpicture}
\begin{axis}[
    scale=0.75,
    xmin=1,
    xmax=10,
    ymin=1500,
    ymax=2000, 
    xlabel={\large Episode},
    ylabel={\large Cost},
    grid=both,
    legend style={
        at={(0.5,-0.25)},
        anchor=north, legend columns=-1
    },
    x=0.8cm, 
    mark size=1.5pt 
]

\pgfplotstableread[col sep=comma]{csv/cost_per_episode_num_samples_50_obstacle.csv}\datatable

\addplot[thick, color=blue, mark=*] 
    table[x=episode, y=original_mppi] {\datatable};
\addplot[thick, color=red, mark=square*] 
    table[x=episode, y=transformer_mppi] {\datatable};

\legend{Original MPPI, Transformer MPPI}
\end{axis}
\end{tikzpicture}
        }
        \caption{Cost vs. episode for 50 samples.}
        \label{fig:cost_vs_episode_navigation}
    \end{subfigure}
    \hfill
    \begin{subfigure}[b]{\subfigwidth}
        \centering
        \resizebox{\textwidth}{!}{
            \begin{tikzpicture}
\begin{axis}[
 scale=0.75,
    ybar,
    ymin=132,
    ymax=145, 
    bar width=0.3cm, 
    legend style={
        at={(0.0,-0.35)},
        anchor=north west,
        legend columns=-1,
        column sep=0.3em 
    },
    ylabel={\large Average steps},
    xlabel={\large Number of samples},
    symbolic x coords={50, 100, 200, 300, 400, 500}, 
    xtick=data,
    x=1.2cm, 
    grid=both 
]

\pgfplotstableread[col sep=comma]{csv/average_steps_vs_num_samples_obstacle.csv}\datatable

\addplot table[x=num_samples, y=original_mppi] {\datatable};
\addplot table[x=num_samples, y=transformer_mppi] {\datatable};

\legend{Original MPPI, Transformer MPPI}
\end{axis}
\end{tikzpicture}
        }
        \caption{Average steps vs. number of samples (over successful episodes).}
        \label{fig:avg_steps_vs_number_of_samples_navigation}
    \end{subfigure}
    \caption{Performance comparison of \ours versus MPPI for navigation 2D}
\end{figure}

Fig. \ref{fig:box_plot_vs_number_of_samples_navigation} provides a box plot that illustrates the cost distribution over 10 successful episodes for this environment, stratified by the number of samples (50, 100, 200, 300, 400, and 500). The comparison between the proposed \ours approach and MPPI reveals that \ours consistently achieves lower median costs for all sample sizes. 
Fig. \ref{fig:avg_cost_vs_number_of_samples_navigation} presents the average cost as a function of sample size, computed over successful episodes. While the average cost for MPPI decreases as the sample size increases, \ours consistently achieves a lower and relatively stable average cost across all sample sizes. This stability indicates the robustness of \ours in generating efficient control sequences regardless of the sampling density.
Fig. \ref{fig:cost_vs_episode_navigation} illustrates the variation in cost across episodes for a scenario with a limited number of samples (i.e., 50). The data reveal a consistent trend in which \ours achieves lower costs compared to MPPI in most episodes. Specifically, in 8 out of the 10 episodes, \ours demonstrates superior performance, achieving reduced costs relative to baseline.
\subsubsection*{Step efficiency}
Fig. \ref{fig:avg_steps_vs_number_of_samples_navigation} compares the average number of steps required to reach the goal state across different sample sizes, averaged over 10 successful episodes. The results reveal that \ours consistently requires fewer steps than MPPI for low number of samples, suggesting more efficient trajectory planning and execution within the obstacle environment.
In Fig. \ref{fig:steps_vs_episode_navigation}, we present the number of steps taken for each episode with fewer samples (i.e 50). Upon examining the figure, it is evident that the number of steps taken varies between \ours and MPPI across episodes. In some episodes, MPPI achieves a lower step count, while in others, \ours takes fewer steps.
\subsubsection*{Efficiency in handling dynamic obstacles}
\begin{figure}[htbp]
    \centering
    \newcommand{\subfigwidth}{0.4\textwidth}
    \newcommand{\subfigheight}{4cm} 
    
    \vspace{0.2cm}
    
    \begin{subfigure}[b]{\subfigwidth}
        \centering
        \resizebox{\textwidth}{!}{
            \begin{tikzpicture}
\begin{axis}[
    scale=0.8,
    xmin=1,
    xmax=10,
    ymin=120,
    ymax=160, 
    xlabel={\large Episode},
    ylabel={\large Steps},
    grid=both,
    legend style={
        at={(0.5,-0.25)},
        anchor=north, legend columns=-1
    },
    x=0.8cm, 
    mark size=1.5pt 
]

\pgfplotstableread[col sep=comma]{csv/steps_per_episode_num_samples_50_obstacle.csv}\datatable

\addplot[thick, color=blue, mark=*] 
    table[x=episode, y=original_mppi] {\datatable};
\addplot[thick, color=red, mark=square*] 
    table[x=episode, y=transformer_mppi] {\datatable};

\legend{Original MPPI, Transformer MPPI}
\end{axis}
\end{tikzpicture}
        }
        \caption{Number of steps vs. episode.}
        \label{fig:steps_vs_episode_navigation}
    \end{subfigure}
    \hfill
    \begin{subfigure}[b]{\subfigwidth}
        \centering
        \resizebox{\textwidth}{!}{
\begin{tikzpicture}
\begin{axis}[
    width=6cm,                
    height=4.5cm,              
    scale only axis,                    
    ymin=1650,
    ymax=1850,                          
    xlabel={\large Dynamic obstacles},
    ylabel={\large Average cost},
    grid=both,
    legend style={
        at={(0.5,-0.25)},              
        anchor=north, 
        legend columns=-1
    },
    mark size=1.5pt,                    
    enlargelimits=false                 
]

    \pgfplotstableread[col sep=comma]{csv/average_cost_vs_num_dynamic_obstacle.csv}\datatable

    \addplot[thick, color=blue, mark=*] 
        table[x=num_obstacles, y=original_mppi] {\datatable};
    \addplot[thick, color=red, mark=square*] 
        table[x=num_obstacles, y=transformer_mppi] {\datatable};

    \legend{Original MPPI, Transformer MPPI}
\end{axis}
\end{tikzpicture}
        }
        \caption{Average cost vs. number of dynamic obstacles (over successful episodes).}
        \label{fig:avg_cost_vs_dynamic_obstacles_navigation}
    \end{subfigure}
     \caption{Performance comparison of \ours versus MPPI for navigation 2D}
    \label{fig:metrics_navigation_2}
\end{figure}
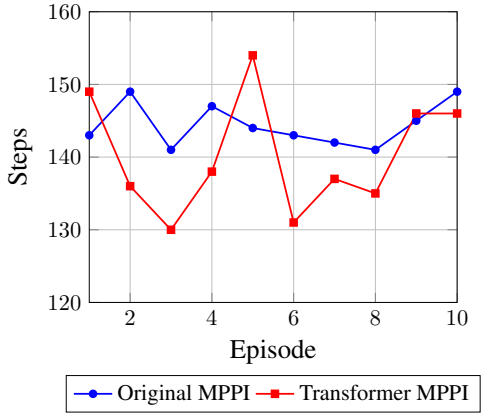
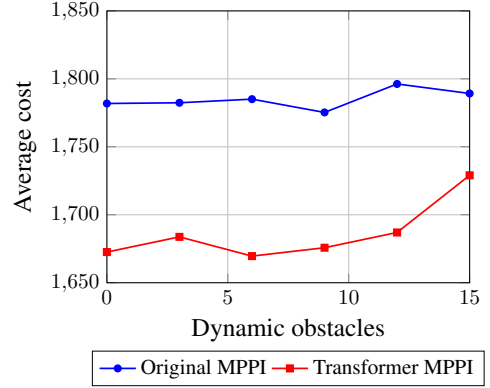

Fig. \ref{fig:avg_cost_vs_dynamic_obstacles_navigation} examines the average cost as a function of the number of dynamic obstacles when the number of samples is 50. Note that this average is taken over 10 successful episodes. In this experiment, the speed of the dynamic obstacles was varied between $0.1$ and $0.5$ $m/s$ to simulate a realistic and challenging environment. As the number of dynamic obstacles increases, both \ours and MPPI experience a slight increase in average cost due to the added complexity of avoiding collisions and navigating efficiently. However, \ours consistently achieves a lower average cost across all levels of obstacle density. This demonstrates the robustness of the transformer-based initialization, which allows \ours to adapt effectively to dynamic obstacle scenarios even when the transformer was not trained on dynamic obstacle configurations. 
\subsection{Autonomous racing\label{subsec:racing}}
\label{subsec:results_racing}
The autonomous racing environment simulates the task of navigating a race track, where the primary objective is to reach the terminal state efficiently while adhering to lane boundaries. The setup is inspired by the virtual racing environment described in \cite{williams2018information}, featuring dynamic track layouts and constraints that challenge the control algorithms to balance speed, stability, and adherence to the track. The agent (a simulated car) operates in a continuous control space and must optimize its trajectory to minimize a predefined cost function while maintaining its position within the track boundaries. This task emphasizes high-speed precision, penalizing deviations from the track center and excessive drifting, both of which can compromise the stability of the car and race performance. The control space for the car consists of two key variables: throttle/brake and steering angle. The cost function was defined as $J(\bx_k)=\sum_{i=k}^{k+N-1}R(\bx_i)$ where $ R(\bx_k) = 2|v_k| - |d| - 5000R^\star_\beta - 1000000R^\star_k $, where $ v_k $ is the car's velocity at time step $k$, $d$ is the car's distance from the track lane's center, $R^\star_\beta$ is an indicator variable for exceeding a certain drift amount $\beta$, and $R^\star_k$ is an indicator variable for when the car leaves the track boundary. Note that $ \beta $ represents the angle between the velocity vector and the longitudinal axis, with a limit set to $45^\circ$.
Note that in the autonomous racing, one of the primary objectives is to maintain a specified speed while navigating the track. This requirement imposes additional constraints on the control process, making lower sample sizes unsuitable due to their tendency to result in frequent collisions. Consequently, the results for this environment start with a minimum of 5000 samples, progressively increasing the sample size from that point to ensure meaningful comparisons.
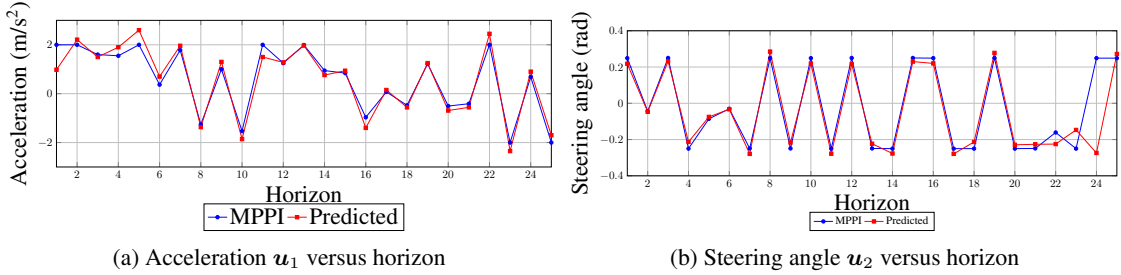
\begin{figure}[htbp]
    \centering

    \newcommand{\subfigwidth}{0.45\textwidth}
    \newcommand{\subfigheight}{6cm} 
    
    \begin{subfigure}[b]{\subfigwidth}
        \centering
        \resizebox{\textwidth}{!}{
            \begin{tikzpicture}
\begin{axis}[
    scale=0.8,
    xmin=1,
    xmax=25,
    ymin=-3,
    ymax=3, 
    xlabel={\huge Horizon},
    ylabel={\huge Acceleration (m/s\textsuperscript{2})},
    grid=both,
    legend style={
    font=\huge,
        at={(0.5,-0.25)},
        anchor=north, legend columns=-1
    },
    x=0.8cm, 
    mark size=1.5pt 
]

\pgfplotstableread[col sep=comma]{csv/transformer_predictions_racing_u1.csv}\datatable

\addplot[thick, color=blue, mark=*] 
    table[x=horizon, y= original] {\datatable};
\addplot[thick, color=red, mark=square*] 
    table[x=horizon, y=transformer] {\datatable};

\legend{MPPI, Predicted}
\end{axis}
\end{tikzpicture}
        }
        \caption{Acceleration $\bu_1$ versus horizon}
        \label{fig:subfig5}
    \end{subfigure}
    \begin{subfigure}[b]{\subfigwidth}
        \centering
        \resizebox{\textwidth}{!}{
            \begin{tikzpicture}
\begin{axis}[
    scale=0.8,
    xmin=1,
    xmax=25,
    ymin=-0.4,
    ymax=0.4, 
    xlabel={\huge Horizon},
    ylabel={\huge Steering angle (rad)},
    grid=both,
    legend style={
        at={(0.5,-0.25)},
        anchor=north, legend columns=-1
    },
    x=0.8cm, 
    mark size=1.5pt 
]

\pgfplotstableread[col sep=comma]{csv/transformer_predictions_racing_u2.csv}\datatable

\addplot[thick, color=blue, mark=*] 
    table[x=horizon, y= original] {\datatable};
\addplot[thick, color=red, mark=square*] 
    table[x=horizon, y=transformer] {\datatable};

\legend{MPPI, Predicted}
\end{axis}
\end{tikzpicture}
        }
        \caption{Steering angle $\bu_2$ versus horizon}
        \label{fig:transformer_predictions_racing_u2}
    \end{subfigure}
    \caption{Transformer predictions for a random sample from test data for autonomous racing}
    \label{fig:transformer_predictions_autonomous_racing}
\end{figure}

\begin{figure}[htbp]
    \centering
    \newcommand{\subfigwidth}{0.45\textwidth}
    \newcommand{\subfigheight}{6cm} 
    
    \begin{subfigure}[b]{\subfigwidth}
        \centering
        \resizebox{\textwidth}{!}{
        \includegraphics[width=0.9\linewidth]{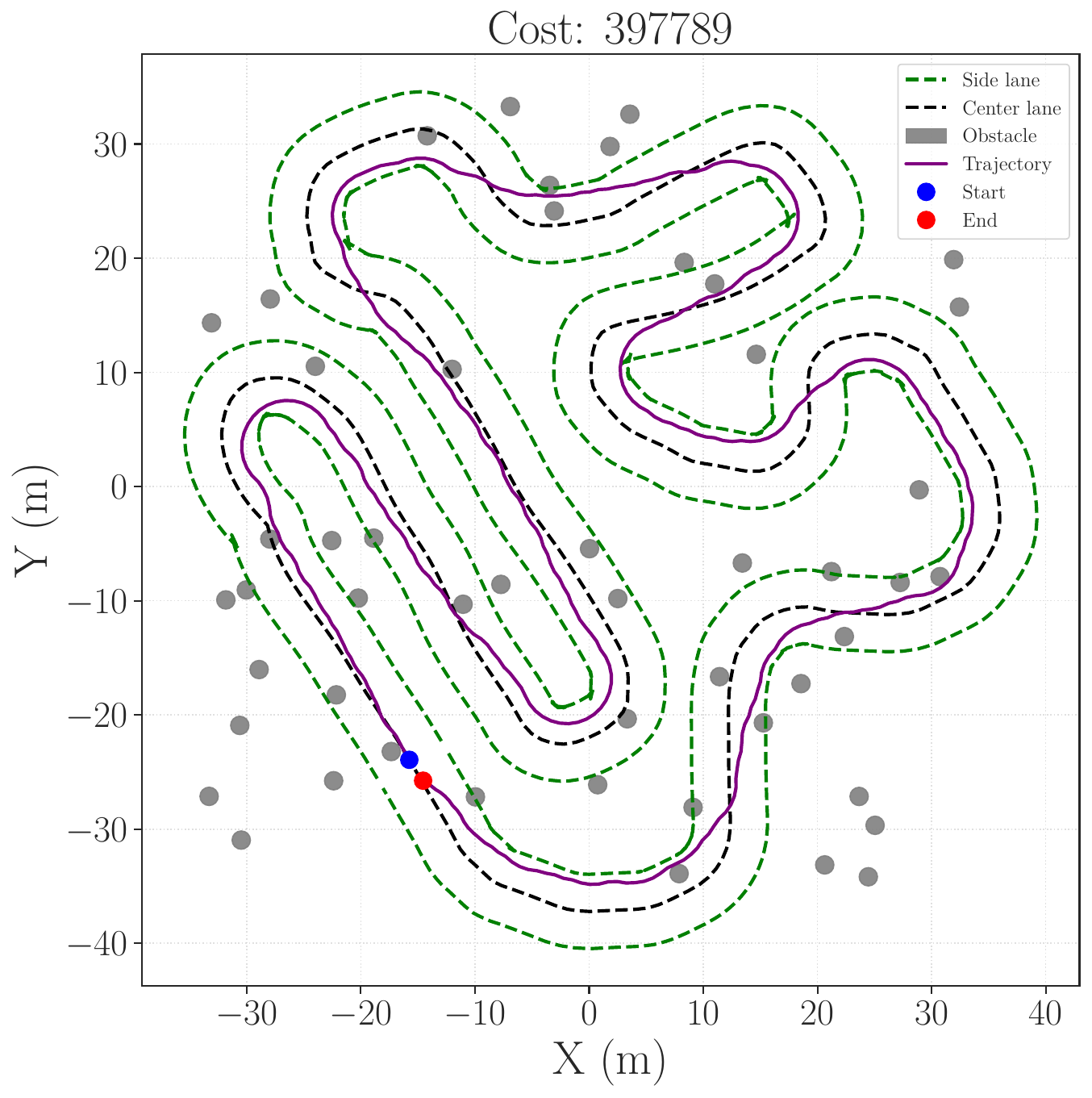}
        }
        \caption{Original MPPI}
    \end{subfigure}
    \begin{subfigure}[b]{\subfigwidth}
        \centering
        \resizebox{\textwidth}{!}{
            \includegraphics[width = 0.9\linewidth]{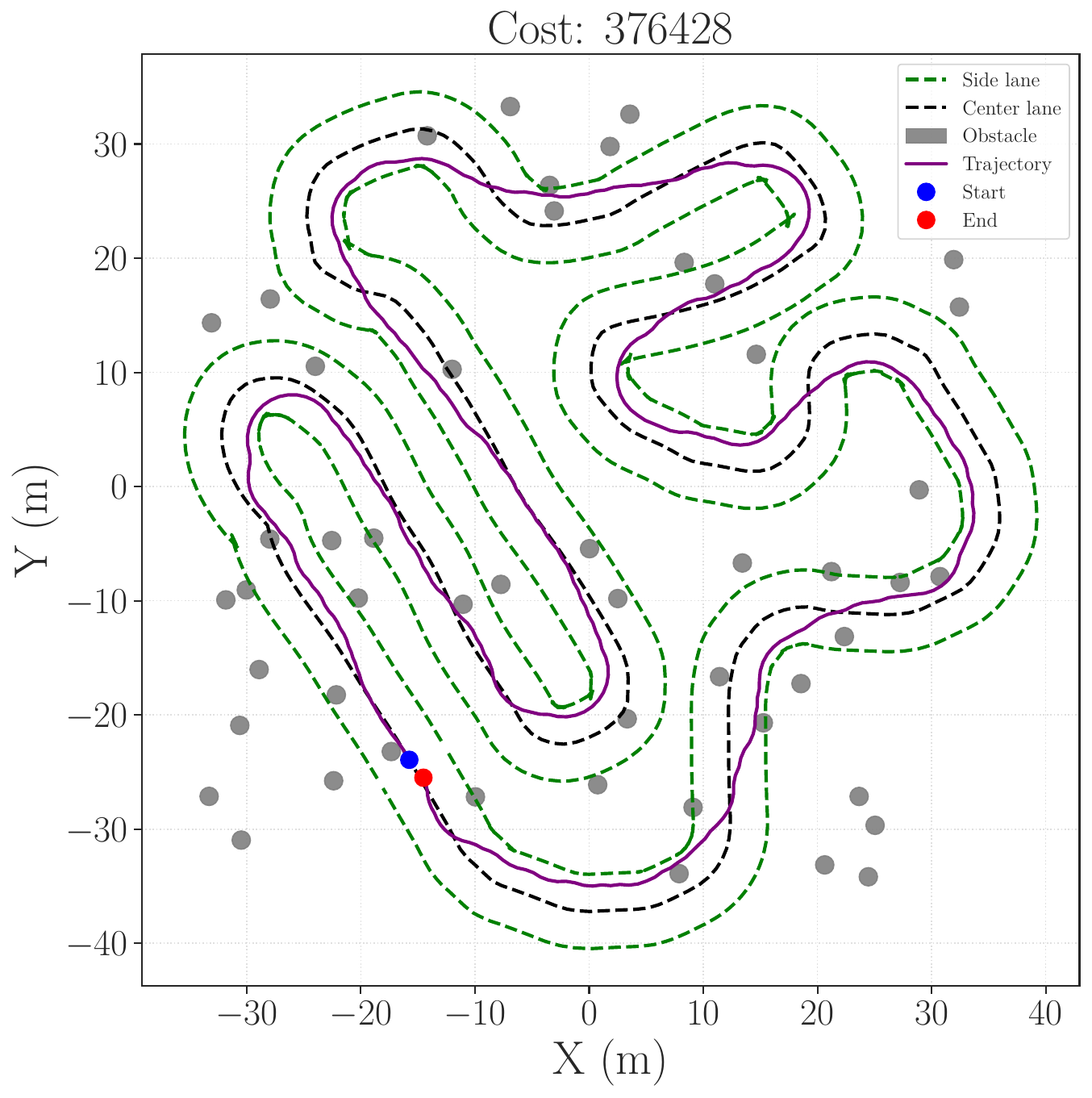}
        }
        \caption{Transformer MPPI}
    \end{subfigure}
    
    \caption{Trajectories generated via MPPI and \ours for autonomous racing when the number of samples is 5000.
    }
    \label{fig:trajectory_racing}
\end{figure}
Fig. \ref{fig:trajectory_racing} compares the trajectories taken by MPPI and \ours in the presence of 50 obstacles for 5000 samples. We can see that the \ours has a relatively lower costs as compared to standard MPPI.
\subsubsection*{Cost analysis}
Fig. \ref{fig:box_plot_vs_number_of_samples_racing} depicts a box plot comparing the cost distribution across 10 episodes for the autonomous racing environment for various sample sizes (i.e., 5000, 6000, 7000, 8000, 9000, and 10000), considering only successful runs. The results indicate that \ours consistently outperforms MPPI, achieving relatively lower median costs across all sample configurations. 
\begin{figure}[htbp]
  \centering
    \newcommand{\subfigwidth}{0.4\textwidth}
    \newcommand{\subfigheight}{4cm} 

    \begin{subfigure}[b]{\subfigwidth}
        \centering
        \resizebox{\textwidth}{!}{
\pgfplotsset{
    boxplot legend/.style={
        legend image code/.code={
            \draw[fill=#1, opacity = 0.6, draw=black] (0cm,-0.2cm) rectangle (0.2cm,0.2cm); 
            \draw[black] (0cm,0cm) -- (0cm,0.1cm); 
            \draw[black] (0.2cm,0cm) -- (0.2cm,0.1cm); 
            \draw[black] (0cm,-0.05cm) -- (0.2cm,-0.05cm); 
            \draw[black] (0.01cm,0.3cm) -- (0.19cm,0.3cm); 
            \draw[black] (0.1cm,0.1cm) -- (0.1cm,0.3cm); 
        }
    }
}
\begin{tikzpicture}
    \pgfplotstableread[col sep=comma]{data_mppi_racing.csv}\csvdataone
    \pgfplotstableread[col sep=comma]{data_transformermppi_racing.csv}\csvdatatwo

    \begin{axis}[
        boxplot/draw direction = y,
        axis x line=bottom, 
        axis y line=left,   
        enlarge y limits,
        ymajorgrids,
        xlabel = {\large Number of samples},
        xtick = {2,5,8,11,14,17}, 
        xticklabel style = {align=center, font=\small, rotate=60},
        xticklabels = {5000, 6000, 7000, 8000, 9000, 10000},
        xtick style = {draw=none}, 
        ylabel = {\large Cost},
        ytick = {360000.00, 370000.00, 380000.00, 390000.00, 400000.00, 410000.00},
        xmin=0, xmax=19,
        clip=false, 
        legend style={
            at={(0.5,-0.3)}, 
            anchor=north,
            legend columns=2, 
            every even column/.append style={column sep=0.4cm}, 
        },
        legend cell align={left}, 
    ]
        \pgfmathsetmacro{\shift}{0.6}
        
        \foreach \n in {0,...,5} { 
            \pgfmathtruncatemacro{\sample}{\n +1} 
            \pgfmathsetmacro{\xcenter}{(\sample-1)*3 +2} 
                
            \addplot+[
                boxplot,
                fill=blue,
                opacity=0.6,
                draw=black,
                boxplot/draw position={\xcenter - \shift},
                forget plot, 
            ] table[y index=\n] {\csvdataone};
            
            \addplot+[
                boxplot,
                fill=red,
                opacity = 0.6,
                draw=black,
                boxplot/draw position={\xcenter + \shift},
                forget plot, 
            ] table[y index=\n] {\csvdatatwo};
        }
        
        \addlegendimage{boxplot legend=blue}
        \addlegendentry{Original MPPI}

        \addlegendimage{boxplot legend=red}
        \addlegendentry{Transformer MPPI}
    \end{axis}
\end{tikzpicture}
        }
        \caption{Box plot of cost vs number of samples (over successful episodes).}
        \label{fig:box_plot_vs_number_of_samples_racing}
    \end{subfigure}
    \hfill
    \begin{subfigure}[b]{\subfigwidth}
        \centering
        \resizebox{\textwidth}{!}{
            \begin{tikzpicture}
\begin{axis}[
    scale=0.8,
    ybar,
    ymin=360000,
    ymax=410000, 
    bar width=0.3cm, 
    legend style={
        at={(0.1,-0.35)},
        anchor=north west,
        legend columns=-1,
        column sep=0.3em 
    },
    ylabel={\large Average cost},
    xlabel={\large Number of samples},
    symbolic x coords={5000, 6000, 7000, 8000, 9000, 10000}, 
    xtick=data,
    x=1.2cm, 
    grid=both, 
]

\pgfplotstableread[col sep=comma]{csv/average_cost_vs_num_samples_racing.csv}\datatable

\addplot[
    fill=blue!50, 
] 
table[x=num_samples, y=original_mppi] {\datatable};

\addplot[
    fill=red!50, 
] 
table[x=num_samples, y=transformer_mppi] {\datatable};

\legend{Original MPPI, Transformer MPPI}
\end{axis}
\end{tikzpicture}
        }
        \caption{Average cost vs. number of samples (over successful episodes).}
        \label{fig:avg_cost_vs_number_of_samples_racing}
    \end{subfigure}
    \caption{Comparison of \ours versus MPPI for autonomous racing}
\end{figure}
Fig. \ref{fig:avg_cost_vs_number_of_samples_racing} illustrates the average cost as a function of the number of samples, averaged over successful episodes for this environment. The results show that \ours consistently achieves a lower average cost across all sample sizes compared to MPPI. However, as the number of samples increases significantly (i.e., approaching 10,000), the costs for both approaches converge to approximately the same value. This convergence indicates that a sufficiently large number of samples can mitigate the differences in initialization strategies, although \ours remains more efficient at lower sample sizes.
The variation in cost across episodes for this environment, using a fixed sample size of 5000, reveals a clear trend where \ours has consistently lower costs compared to MPPI, as shown in Fig. \ref{fig:cost_vs_episode_racing}.
\begin{figure}[htbp]
  \centering
    \newcommand{\subfigwidth}{0.4\textwidth}
    \newcommand{\subfigheight}{4cm} 

        \begin{subfigure}[b]{\subfigwidth}
        \centering
        \resizebox{\textwidth}{!}{
            \begin{tikzpicture}
\begin{axis}[
    scale=0.75,
    xmin=1,
    xmax=10,
    ymin=360000,
    ymax=410000, 
    xlabel={\large Episode},
    ylabel={\large Cost},
    grid=both,
    legend style={
        at={(0.5,-0.25)},
        anchor=north, legend columns=-1
    },
    x=0.8cm, 
    mark size=1.5pt 
]

\pgfplotstableread[col sep=comma]{csv/cost_per_episode_num_samples_50_racing.csv}\datatable

\addplot[thick, color=blue, mark=*] 
    table[x=episode, y=original_mppi] {\datatable};
\addplot[thick, color=red, mark=square*] 
    table[x=episode, y=transformer_mppi] {\datatable};

\legend{Original MPPI, Transformer MPPI}
\end{axis}
\end{tikzpicture}
        }
        \caption{Cost vs. episode for 5000 samples.}
        \label{fig:cost_vs_episode_racing}
    \end{subfigure}
    \hfill
    \begin{subfigure}[b]{\subfigwidth}
        \centering
        \resizebox{\textwidth}{!}{
            \begin{tikzpicture}
\begin{axis}[
 scale=0.75,
    ybar,
    ymin=415,
    ymax=430, 
    bar width=0.3cm, 
    legend style={
        at={(0.0,-0.35)},
        anchor=north west,
        legend columns=-1,
        column sep=0.3em 
    },
    ylabel={\large Average steps},
    xlabel={\large Number of samples},
    symbolic x coords={5000, 6000, 7000, 8000, 9000, 10000}, 
    xtick=data,
    x=1.2cm, 
    grid=both 
]

\pgfplotstableread[col sep=comma]{csv/average_steps_vs_num_samples_racing.csv}\datatable

\addplot table[x=num_samples, y=original_mppi] {\datatable};
\addplot table[x=num_samples, y=transformer_mppi] {\datatable};

\legend{Original MPPI, Transformer MPPI}
\end{axis}
\end{tikzpicture}
        }
        \caption{Average steps vs. number of samples (over successful episodes). }
        \label{fig:avg_steps_vs_number_of_samples_racing}
    \end{subfigure}
     \begin{subfigure}[b]{\subfigwidth}
        \centering
        \resizebox{\textwidth}{!}{
            \begin{tikzpicture}
\begin{axis}[
    scale=0.8,
    xmin=1,
    xmax=10,
    ymin=410,
    ymax=450, 
    xlabel={\large Episode},
    ylabel={\large Steps},
    grid=both,
    legend style={
        at={(0.5,-0.25)},
        anchor=north, legend columns=-1
    },
    x=0.8cm, 
    mark size=1.5pt 
]

\pgfplotstableread[col sep=comma]{csv/steps_per_episode_num_samples_50_racing.csv}\datatable

\addplot[thick, color=blue, mark=*] 
    table[x=episode, y=original_mppi] {\datatable};
\addplot[thick, color=red, mark=square*] 
    table[x=episode, y=transformer_mppi] {\datatable};

\legend{Original MPPI, Transformer MPPI}
\end{axis}
\end{tikzpicture}
        }
        \caption{Number of steps vs. episode for 5000 samples}
        \label{fig:steps_vs_episode_racing}
    \end{subfigure}
    \hfill
    \begin{subfigure}[b]{\subfigwidth}
        \centering
        \resizebox{\textwidth}{!}{
\begin{tikzpicture}
\begin{axis}[
    width=6cm,                
    height=4.5cm,              
    scale only axis,                    
    ymin=370000,
    ymax=410000,                          
    xlabel={\large Dynamic obstacles},
    ylabel={\large Average cost},
    grid=both,
    legend style={
        at={(0.5,-0.25)},              
        anchor=north, 
        legend columns=-1
    },
    mark size=1.5pt,                    
    enlargelimits=false                 
]

    \pgfplotstableread[col sep=comma]{csv/average_cost_vs_num_dynamic_racing.csv}\datatable

    \addplot[thick, color=blue, mark=*] 
        table[x=num_obstacles, y=original_mppi] {\datatable};
    \addplot[thick, color=red, mark=square*] 
        table[x=num_obstacles, y=transformer_mppi] {\datatable};

    \legend{Original MPPI, Transformer MPPI}
\end{axis}
\end{tikzpicture}
        }
        \caption{Average cost vs. number of dynamic obstacles (over successful episodes).}
        \label{fig:avg_cost_vs_dynamic_obstacles_racing}
    \end{subfigure}
    \caption{Comparison of \ours versus MPPI for autonomous racing}
\end{figure}
\subsubsection*{Step efficiency}
Fig. \ref{fig:steps_vs_episode_racing} shows the step taken by the agent in different episodes for 5000 samples, while the average number of steps as a function of the number of samples is plotted in Fig. \ref{fig:avg_steps_vs_number_of_samples_racing}. No clear trend is observed here, as both \ours and MPPI take a similar number of steps to reach the goal across all sample sizes.
\subsubsection*{Efficiency in handling dynamic obstacles}
Fig. \ref{fig:avg_cost_vs_dynamic_obstacles_racing} presents the average cost as a function of the number of dynamic obstacles, averaged over successful episodes, with obstacle velocities varying again between $0.1$ and $0.5$ $m/s$. As obstacle density increases, the environment becomes more challenging, resulting in a slight increase in average cost for both \ours and MPPI. However, \ours shows a clear advantage in consistently achieving lower costs across all obstacles levels.

\section{Conclusions}
\label{sec:conclusions}

In this paper, we address the challenges of sample efficiency and computational costs in MPPI control by introducing \ours, a transformer-based approach for initializing the mean control sequence. Our method reduces the computational burden and accelerates convergence by leveraging the ability of transformer to generate informed initial control sequences based on historical optimal control data. We synthesized optimal control inputs by integrating the predictions of transformer with the MPPI framework and validated the efficacy of our approach on control tasks, such as collision avoidance in 2D environments and autonomous racing amid static and dynamic obstacles. The results demonstrated significant improvements in overall average cost, sample efficiency and computational speed compared to traditional MPPI algorithms, highlighting the feasibility of \ours for real-time applications.
Future work includes deploying \ours in physical robotic systems to assess their performance in real world scenarios. Additionally, investigating its applicability in multi-agent systems could broaden its scope of application.

\bibliographystyle{unsrtnat}
\bibliography{references}

\end{document}